\documentclass{article}
\usepackage{amssymb}
\usepackage{graphicx}
\usepackage{xcolor}
\usepackage{booktabs}
\usepackage{makecell}  % Add this in your preamble
\usepackage{amsmath} 
\usepackage{hyperref}
\usepackage{wrapfig} % in your preamble
\usepackage{mathrsfs}
\usepackage{pifont} % Add this in your preamble
\usepackage{multirow}   % In preamble
\usepackage{arydshln}   % In preamble for dashed lines
\usepackage{paralist}
\usepackage{enumitem}

\usepackage[preprint]{corl_2025} % Uncomment for pre-prints (e.g., arxiv); This is like ``final'', but will remove the CORL footnote.

\title{CLASS: Contrastive Learning via Action Sequence Supervision for Robot Manipulation
}

\author{
  Sung-Wook Lee, Xuhui Kang, Brandon Yang, Yen-Ling Kuo\\
  University of Virginia\\
  \texttt{\{dcs3zc,xuhui,branyang,ylkuo\}@virginia.edu}
  \vspace{-14pt}
  \\
}

\begin{document}
\maketitle

%===============================================================================

\begin{abstract}
Recent advances in Behavior Cloning (BC) have led to strong performance in robotic manipulation, driven by expressive models, sequence modeling of actions, and large-scale demonstration data. However, BC faces significant challenges when applied to heterogeneous datasets, such as visual shift with different camera poses or object appearances, where performance degrades despite the benefits of learning at scale. This stems from BC's tendency to overfit individual demonstrations rather than capture shared structure, limiting generalization. To address this, we introduce Contrastive Learning via Action Sequence Supervision (CLASS), a method for learning behavioral representations from demonstrations using supervised contrastive learning. CLASS leverages weak supervision from similar action sequences identified via Dynamic Time Warping (DTW) and optimizes a soft InfoNCE loss with similarity-weighted positive pairs. 
We evaluate CLASS on 5 simulation benchmarks and 3 real-world tasks to achieve competitive results using retrieval-based control with representations only. Most notably, for downstream policy learning under significant visual shifts, Diffusion Policy with CLASS pre-training achieves an average success rate of \(75\%\), while all other baseline methods fail to perform competitively.

% demonstrating the robustness of the learned representations in heterogeneous settings.

% Notably, CLASS relies solely on representation learning and search-based inference, eliminating the need for a policy head, making it well-suited for high-frequency robot control. For representation-based evaluations, CLASS outperforms all the other methods across the tasks. Furthermore, in heterogeneous data collection setups, CLASS even significancy outperforms BC with a policy head.
\end{abstract}

% Two or three meaningful keywords should be added here
\keywords{Robot Manipulation, Action Chunking, Supervised Contrastive Learning, Vision Representation Learning
\vspace{-2pt}
} 
%===============================================================================
% Submission to CoRL 2025 will be entirely electronic, via a web site (not email). Information about the submission process and \LaTeX{} templates are available on the conference web site at \url{https://corl.org/}. For camera ready submission, use the \texttt{final} option for the \verb|\usepackage| command. 

\section{Introduction}
\vspace{-2pt}

Behavior Cloning (BC) has demonstrated strong performance in robotic manipulation by leveraging expressive models and action sequence modeling. Efforts to improve BC have focused on large-scale dataset collection \citep{brohan2022rt, khazatsky2024droid} and advances in model architectures \citep{chi2024diffusionpolicy, zhao2023learning, lee2024behavior} to better capture the complex distribution of demonstration data. However, expressive policies often struggle to generalize, especially when trained on demonstrations collected under heterogeneous conditions—that is, where the policy must adapt to additional properties not present in homogeneous data, such as changes in viewpoint or object appearance \citep{xie2024decomposing, saxena2025matters}. This suggests a tendency to overfit individual actions and a limited ability to capture shared structure across demonstrations \citep{li2025beyond}.

To address this, we propose \textbf{Contrastive Learning via Action Sequence Supervision (CLASS)}, a framework for learning behaviorally grounded representations from demonstrations using supervised contrastive learning. Rather than relying on direct action prediction, CLASS supervises the encoder by aligning observations based on action sequence similarity, measured via Dynamic Time Warping (DTW), encouraging states that lead to similar future behaviors to cluster in the latent space. This weak supervision enables the model to capture shared structure across demonstrations, improving robustness to variations in visual conditions such as camera pose and object appearance. The learned representation supports both retrieval-based inference and policy fine-tuning, and consistently improves performance across both homogeneous and heterogeneous data settings. Across a range of simulated and real-world robotic manipulation tasks, CLASS achieves strong gains over behavior cloning and representation learning baselines, demonstrating its ability to learn more transferable and composable behavioral representations.

Our contributions are as follows:
1) We introduce CLASS, a method for learning robust visual representations from demonstrations by aligning observations based on action sequence similarity.
2) We develop a novel scheme for soft contrastive learning that focuses on weighting similar action sequences according to their DTW-based similarity.
3) We show that CLASS outperforms behavior cloning and representation learning baselines across a range of simulated and real-world robotic manipulation tasks, particularly under heterogeneous visual conditions.

%===============================================================================

\begin{figure*}[t]
    \begin{center}
        {\includegraphics[width=0.999\linewidth]{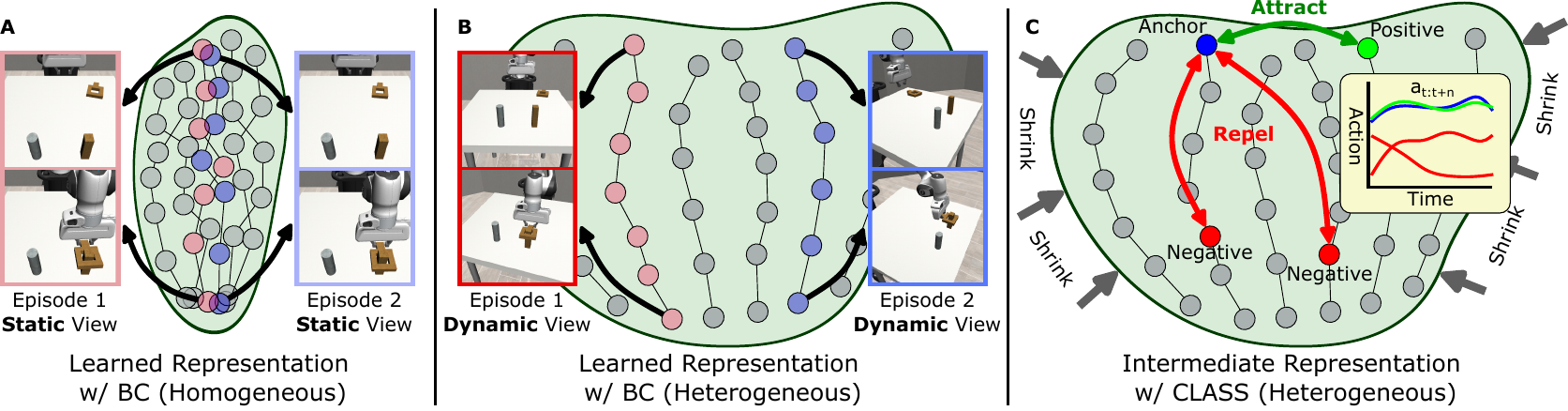} 
        }
        \vspace{-12pt}
        \caption{\textbf{Comparison between Behavior Cloning (BC) and Contrastive Learning via Action Sequence Supervision (CLASS).}  
\textbf{(A)} With homogeneous demonstrations with consistent visual conditions, BC learns a compact representation with high transferability.
\textbf{(B)} With heterogeneous demonstrations such as varying viewpoints, BC overfits to individual state-action pairs and generalizes poorly.
\textbf{(C)} CLASS addresses this by attracting states with similar action sequences and repelling those with dissimilar ones, with a soft supervised contrastive learning objective to learn more robust and composable representations.}
    \end{center}
    \label{fig:main}%
\vspace{-14pt}
\end{figure*}
%===============================================================================
\vspace{-4pt}
\section{Related Work}
\vspace{-2pt}

\textbf{Action Sequence Modeling (Action Chunking)}
Real-world physical tasks demand executing action sequences that are both temporally coherent and sensitive to system dynamics. However, traditional policy learning methods that rely on the Markov assumption and produce actions one step at a time have struggled with long-horizon tasks with compounding errors \cite{NEURIPS2021_099fe6b0, lee2024diff}.
Recent research addresses these issues by casting policy learning as a sequence modeling problem, an approach that has demonstrated strong empirical performance across domains \citep{lai2022action, janner2021offline}. For behavior cloning in robot manipulation, Diffusion Policy (DP)~\citep{chi2024diffusionpolicy} incorporates a diffusion framework with action chunking to achieve the state-of-the-art performance. Action Chunking Transformer~\citep{zhao2023learning} uses temporal ensembling between previously generated actions to further improve temporal consistency. In addition, VQ-BET~\citep{lee2024behavior} leverages Vector Quantization \citep{NIPS2017_7a98af17} to discretize continuous action chunks. These discrete representations are then integrated into the Behavior Transformer \citep{shafiullah2022behavior}, leading to improvements in both inference speed and policy performance. While the previous action chunking methods rely on strong supervision via behavior cloning, our method focuses on using the similarity between action sequences as a contrastive signal for representation learning.

\vspace{-2pt}
\textbf{Contrastive Representation Learning}
Contrastive learning has been applied across different fields to extract informative representations from high-dimensional inputs. %areas such as reinforcement learning (RL), imitation learning (IL), and trajectory prediction 
In reinforcement learning (RL), it is often used to encourage representations that capture task-relevant dynamics and improve sample efficiency. For example, \citep{eysenbach2022contrastive} contrasts observations based on whether a sampled state is likely to occur in the future relative to a given anchor state.
In imitation learning, time-contrastive objectives~\cite{sermanet2018time,nair2022r3m,ma2022vip} within each demonstration are applied to learn visual representations for downstream robot manipulation tasks.
In trajectory prediction, \cite{halawa2022action} proposes a hierarchical contrastive approach that uses high-level action information to bring similar behaviors closer in the embedding space. In autonomous driving, \cite{zhang2022learning} uses pseudo-labeled actions from videos to identify positive pairs in a binary manner to perform contrastive learning on vision representations.  
More broadly, recent works \citep{zhuang2024not, suresh2021not, li2023rethinking, lee2023soft} explore soft contrastive learning by assigning continuous similarity-based weights, providing finer-grained learning signals that better capture inter-sample relationships.
% In time-series classification, \cite{lee2023soft} incorporates dynamic time warping (DTW) with soft assignments to compute similarity scores between anchors and negatives for contrastive objective. 
Unlike prior contrastive methods in robotics that focus on learning temporal relationships within individual demonstrations, our method composes similar behaviors across different demonstrations. In addition, whereas most previous soft contrastive learning methods focus on assigning similarity-based weights to negative pairs, we focus on soft similarity weighting between positive pairs. This design choice aligns well with the goal of imitation learning and retrieval-based policy learning methods like VINN~\citep{pari2021surprising}, where the quality of the learned representation depends on how well similar demonstrations are clustered.

\vspace{-2pt}
\section{Problem Statement}
\vspace{-2pt}

\newcommand{\vect}[1]{\boldsymbol{\mathbf{#1}}}

We address imitation learning by reformulating it as a behaviorally grounded representation learning problem. The goal is to learn an embedding space where observations with similar future behaviors are mapped close together, while dissimilar ones are separated. Instead of relying on direct action supervision, we propose to learn the embedding space via a soft supervised contrastive objective guided by trajectory-level action similarity.

Specifically, given a dataset of demonstration trajectories consisting of observation and action sequences, we define \( \vect{O}_t = \{ o_{t-T_o-1}, \ldots, o_{t-1}, o_t \} \) and \( \vect{A}_t = \{ a_t, a_{t+1}, \ldots, a_{t+T_p-1} \} \), where \( T_o \) is the number of past observations and \( T_p \) is the number of future actions. We seek to learn an encoder \( f_\theta \) that maps the observation sequence \( \vect{O}_t \) to a latent representation \( z = f_\theta(\vect{O}_t) \). The learned embedding should exhibit two key properties: (i) observation sequences that lead to behaviorally similar trajectories should be mapped close together in the latent space, and (ii) sequences from behaviorally dissimilar trajectories should be mapped farther apart.

% To this end, we employ an encoder \( f_\theta \) that transforms each observation \( o_t \) into a compact latent representation:
% \[
% z_t = f_\theta(o_t).
% \]

% By decoupling representation learning from direct action prediction, this approach enables ().
\vspace{-2pt}
\section{Method}
\vspace{-2pt}

In this section, we describe the components of our method. CLASS consists of two stages: (a) pre-computing pairwise action sequence similarities from demonstration data, and (b) representation learning via supervised contrastive learning with a Soft InfoNCE loss.

\vspace{-3pt}
\subsection{Action Sequence Similarity}
\vspace{-2pt}

The first stage of our approach involves measuring similarity between action sequences using Dynamic Time Warping (DTW)~\cite{muller2007dynamic, aeon24jmlr}, which accounts for temporal misalignment between time-series data. Given two action sequences \( \vect{A}_T^m = \{ a_p^m \}_{p=T}^{T+T_p-1} \) and \( \vect{A}_{T'}^n = \{ a_q^n \}_{q=T'}^{T'+T_q-1} \), where \( a_p^m \) and \( a_p^n \) denote the actions at timesteps \(p\) and \(q\) from demonstration \(m\) and \(n\) in the first and second sequences, respectively, the DTW distance is computed as follows: 
\vspace{-2pt}
\[
\text{DTW}( \vect{A}_T^m, \vect{A}_{T'}^n) = \min_{\Gamma} \sum_{(p, q) \in \Gamma} \left\| a_p^m - a_q^n \right\|_2,
\vspace{-5pt}
\] 
% \vspace{-12pt}
where \( \Gamma \) is a monotonic alignment, and \( \| \cdot \|_2 \) denotes the Euclidean distance.
\begin{wrapfigure}[15]{b}{0.545\linewidth}  % 'r' for right, 0.4\linewidth width
    \centering
    \vspace{-13pt}    
    \includegraphics[width=1.04\linewidth, keepaspectratio]{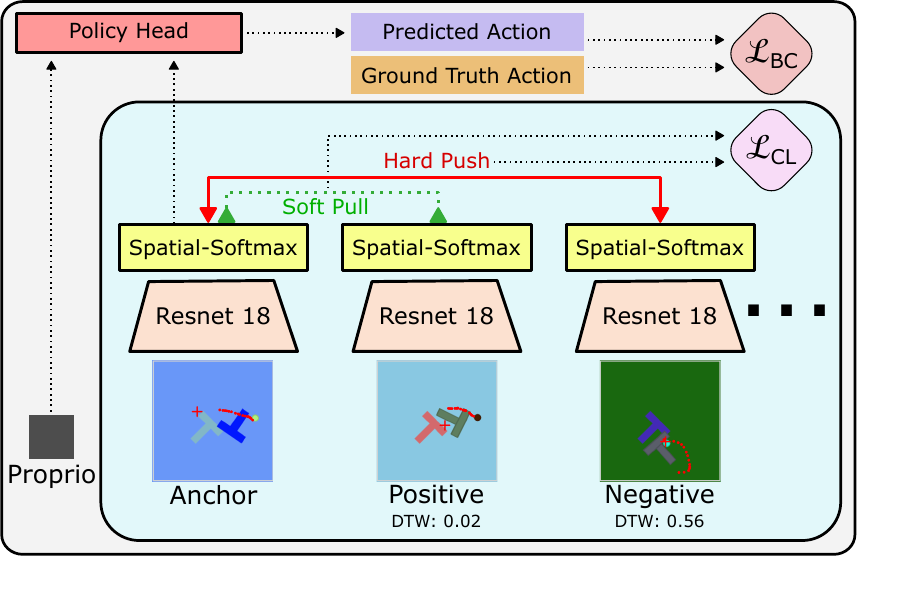}
    \vspace{-28pt}
    \caption{\textbf{Training pipeline}. The inner block corresponds to contrastive representation learning as pre-training. The outer block performs policy fine-tuning with proprioception and the vision representation. Modules with the same colors indicate shared weights.}    
    \label{fig:train}
\end{wrapfigure}
Pairwise DTW distances are computed across all action sequences in the dataset, 
and positive pairs are defined as those with distances below a fixed quantile threshold \( K \). The DTW distances of positive pairs are stored for contrastive learning.

\vspace{-2pt}
\subsection{Soft Contrastive Learning}
\vspace{-2pt}
%Contrastive Loss with Soft Positives via Action Sequence Similarity

To learn behaviorally informed representation, we use a contrastive learning objective that incorporates \textit{soft positive supervision} using the pre-computed DTW values. 
Our method resembles that of \citep{khosla2020supervised, ge2023soft}, with modifications for compatibility with policy learning.
Specifically, given an input image observation batch, each image serves as its own anchor, augmented with random cropping and Gaussian noise. During training, the pre-defined positive pairs, identified via DTW, are pulled together, while all other samples are pushed apart.
The encoder \( f_\theta \) is composed of a ResNet-18 \citep{he2016deep} backbone and a spatial softmax layer, as shown in Fig.~\ref{fig:train}. Given an input batch with data augmentation, we encode it into \( z \) using \( f_\theta \) to obtain the L2-normalized image latents \( \{ \hat{z}_i \}_{i=1}^B \), where \( \hat{z}_i = z_i / \|z_i\|_2 \) and \( \|\hat{z}_i\|_2 = 1 \).
The pairwise similarity matrix \( S \in \mathbb{R}^{B \times B} \) is computed as
\(
S_{ij} = \hat{z}_i^\top \hat{z}_j / \tau %\frac{s_i^\top s_j}{\tau},
\)
where \( \tau > 0 \) is a temperature hyperparameter. We define a soft positive weight \( w_{ij} \in [0, 1] \) between anchor \( i \) and candidate \( j \), computed via DTW-based similarity:
\[
w_{ij} = 
\begin{cases}
1 - \text{CDF}\left(\text{DTW}(i,j)\right), & \text{if $(i,j)$ is a positive pair}\\[0.5em]
0, & \text{otherwise},
\end{cases}
\]
where \(\text{CDF}\) denotes the empirical cumulative distribution function computed over all anchor-positive DTW distances. 
% To control the contribution of negative pairs, we scale negative logits by adding a constant scaling factor \(\rho\) to all negative pairs:
% \[
% \log p_{ij} = S_{ij} - \log \sum_{k=1, k \ne i}^{B} \exp\left(S_{ik} + \rho \mathbb{I}[w_{ik} = 0]\right).
% \]
% \[
% \log p_{ij} = S_{ij} - \log \sum_{k=1, k \ne i}^{B} \exp S_{ik}.
% \]
The soft InfoNCE loss is defined as a weighted log-softmax over positive pairs:
\begin{equation}
\mathcal{L}_{\text{CL}} = -\frac{1}{B}\sum_{i=1}^{B} 
\frac{\sum_{j=1}^{B} w_{ij}\log p_{ij}}{\sum_{j=1}^{B} w_{ij}}, \quad \text{where} \quad  p_{ij} = \frac{ \exp(S_{ij}) }{ \sum_{k=1,\, k \ne i}^{B} \exp(S_{ik}) } 
\label{eq:soft_info_nce}
\end{equation}

Eq.~\eqref{eq:soft_info_nce} generalizes InfoNCE by weighting positives using soft DTW-based scores \( w_{ij} \). This approach enables graded similarity supervision by allowing positive pairs to contribute proportionally to their action sequence similarity. For additional discussion about pre-computation, see Appendix \ref{appx:precompute}.
% Since the loss only requires DTW-based weights for anchor-positive pairs, our method avoids constructing the full \( N \times N \) distance matrix, where $N$ is the number of total data points, significantly reducing memory cost to facilitate learning at scale. 
%For larger datasets, additional savings may be achieved through online similarity computation or further sparsification techniques.

\subsection{Training}
\label{s:train}
We use two training schemes. In the first, the encoder is trained solely with the representation learning objective (e.g., the inner block in Fig.~\ref{fig:train}) without a policy head. 
In the second scheme, we follow a two-stage protocol: the encoder is first pretrained with the representation objective, 
% after which the projection head is discarded 
and the encoder is fine-tuned with a policy head using the behavior cloning (BC) loss. Additional training details are available in Appendix~\ref{appx:implementation}.

\subsection{Policy Evaluation} 
\label{s:eval}
We evaluate policies under two different settings.
In the \textbf{parametric} setting, actions are generated directly by the learned policy network, following standard policy inference procedures.
The parametric evaluation requires training a policy head.
In the \textbf{non-parametric} setting, we extend the approach of \citet{pari2021surprising} to action sequence rollout. At inference time, the most recent observation is encoded into a latent space using the vision encoder, and the \(k_{\text{nn}}\) nearest neighbors are retrieved from the training data using cosine similarity between representations. Their corresponding action sequences are then ensembled using a similarity-weighted average as follows:
\[
\hat{\vect{A}} = \frac{\sum_{i=1}^{k_\text{nn}} \exp\left(c_i\right) \cdot \vect{A}^{(i)}}{\sum_{i=1}^{k_\text{nn}} \exp\left(c_i\right)}, \quad \text{with } c_i = \frac{\text{cos\_sim}(\hat{z}, \hat{z}^{(i)})}{\tau_\text{nn}}
\]
where \(\hat{\vect{A}}\) represents the action sequence rolled out during evaluation, $c_i$ denotes the cosine similarity-based weight, $\hat{z}$ denotes the L2-normalized latent vector obtained by concatenating proprioception and image features, \(\tau_\text{nn}\) is a temperature hyperparameter, and (\(\hat{z}^{(i)}, {\vect{A}^{(i)}}\)) are \(i\)th nearest latent observation and action sequence pair in the dataset.
The non-parametric setting relies solely on the learned representation and does not require a policy head. 

\subsection{Theoretical Analysis}
The proposed method leverages similarity in action sequences to guide representation learning.
As a theoretical justification, we show the relationship between action sequence supervision and the learned latent representation when optimizing the proposed loss function.

\textbf{Proposition 1.} \textit{$\mathcal{L}_{\text{CL}}$ optimizes the KL divergence between the distribution of soft positive weights $w_{ij}$ of actions and the distribution of corresponding similarities in the latent representation.}

See Appendix~\ref{appx:proposition} for the proof.
This proposition establishes the relationship between the similarities in the latent representation space and the sequence-level similarities between action trajectories.
When the representation is learned, the error is bounded by KL plus a constant factor determined by the quantile threshold we set for the positive pairs.

\vspace{-3pt}
\section{Experiments}
\vspace{-2pt}

We conduct extensive experiments in both simulation and real-world tasks. Our investigation addresses the following questions.
\begin{compactenum}
    \item[\textbf{RQ1}] Can CLASS learn representations that achieve competitive retrieval-based policy performance in both homogeneous and heterogeneous settings?
    \item[\textbf{RQ2}] Does the representation learned with CLASS transfer to downstream policy learning through fine-tuning with BC?
    \item[\textbf{RQ3}] How well does CLASS generalize across different settings, including different action spaces and multi-task learning?
\end{compactenum}

\vspace{-2pt}
\subsection{Tasks and Datasets}
\vspace{-2pt}
\label{ss:tasks}
We evaluate CLASS on 5 simulated and 3 real-robot tasks. Each task represents a class of robotic manipulation tasks. See Appendix~\ref{appx:env} for environment details.

\begin{itemize}[nosep,leftmargin=1em,labelwidth=*,align=left]
\item \textbf{Square} is a task from robomimic \citep{mandlekar2021matters}, where the robot places a square nut into a rod, evaluating the basic pick-and-place skill.
We use 200 proficient human demonstrations.

\item \textbf{Three-Stack} is a high-equivariance stacking task from MimicGen \citep{mandlekar2023mimicgen}, where the robot stacks three colored blocks in a specified order.
The dataset contains 1000 expert demonstrations with the default reset distribution.

\item \textbf{Aloha-Transfer} \citep{zhao2023learning} involves a bi-dexterous manipulation task rendered from a top-down view using the Aloha robot.
The dataset includes 100 demonstrations.
% , comprising 50 scripted and 50 human-executed episodes. 
Unlike the other manipulation tasks, it uses bi-robot joint position control, allowing us to assess if CLASS generalizes to different action spaces.

\item \textbf{LIBERO-Object} \citep{liu2023libero} is a multi-task benchmark consisting of ten object-centric pick-and-place tasks, each paired with a natural language instruction encoded using a pretrained BERT \citep{devlin2019bert}. Each task includes 50 human-teleoperated demonstrations.

\item \textbf{Push-T} \citep{chi2024diffusionpolicy} requires the robot to push a T-shaped object into a designated goal on a 2D plane, evaluating a contact-rich setting. A rollout is considered successful if the intersection-over-union between the object and the goal region exceeds 90\%.
The dataset has 206 human demonstrations.

\item \textbf{Two-Stack} is a real-world equivariant task where a red cube is stacked on top of a green cube. 200 demonstrations are collected.

\item \textbf{Mug-Hang} is a real-world precise placement task where the robot picks up a mug placed at a random pose and hangs it onto a mug rack. 200 demonstrations are collected.

\item \textbf{Toaster-Load} is a real-world task where the robot picks up a piece of soft bread and places it into a toaster slot at a random pose. 200 demonstrations are collected.
\end{itemize}

\begin{figure*}[t]
    \begin{center}
        \includegraphics[width=0.72\linewidth]{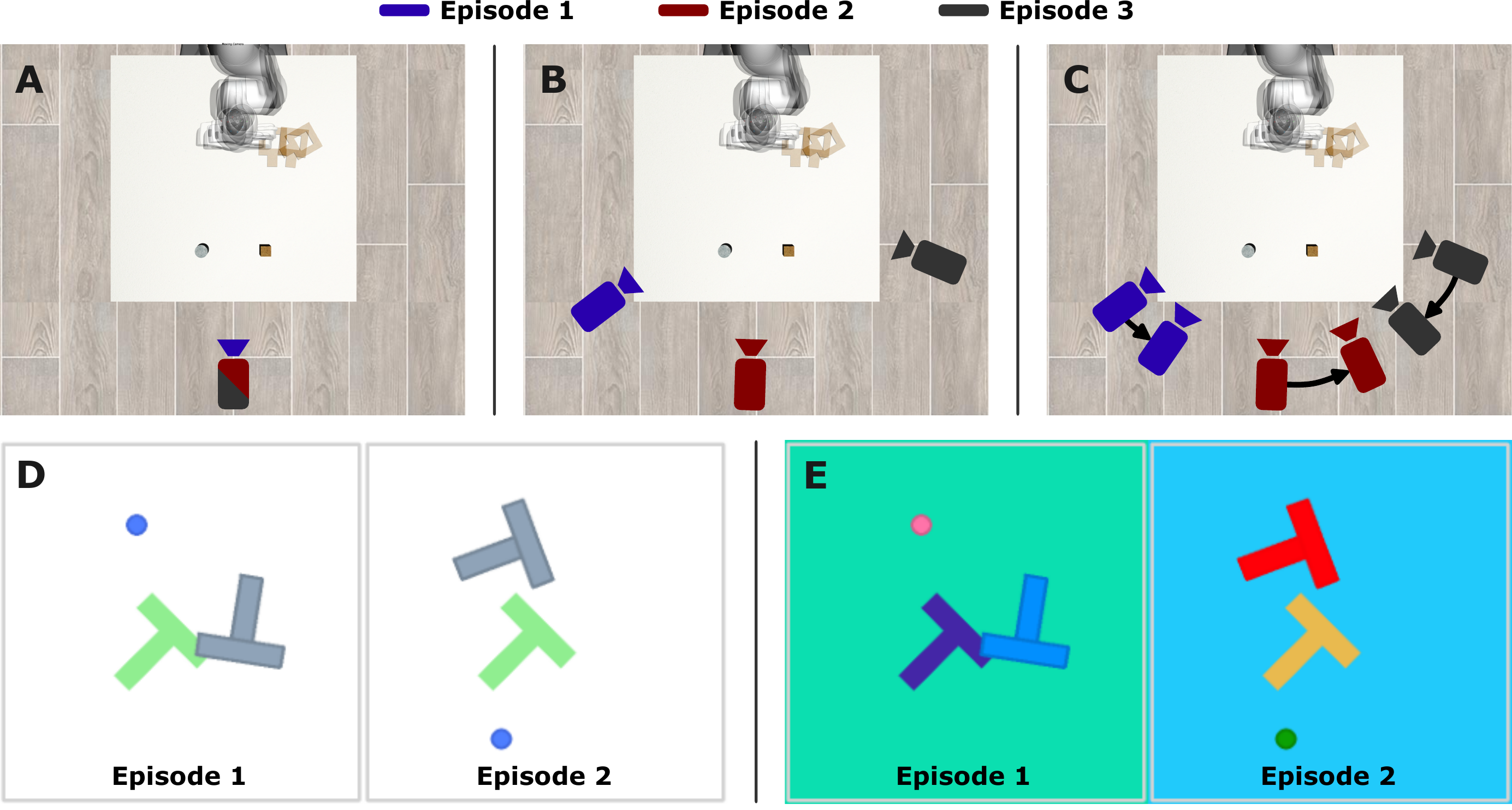}
\vspace{-6pt}
\caption{\textbf{Heterogeneous data collection setups.} 
\textbf{(A)} Fixed camera (\textbf{Fixed-Cam}), commonly used in conventional behavior cloning pipelines. 
\textbf{(B)} Random static camera (\textbf{Rand-Cam}), where the camera pose is randomly sampled at the start of each episode but remains fixed during the episode.
\textbf{(C)} Dynamic camera (\textbf{Dyn-Cam}), where a randomly initialized camera moves during the episode with a random direction while maintaining a consistent look-at target.
\textbf{(D)} Fixed object color (\textbf{Fixed-Color}), commonly assumed in vision-based behavior cloning tasks.
\textbf{(E)} Random object color (\textbf{Rand-Color}), where the color of the objects is randomly varied in each demonstration. We compare setups (A) and (C) for simulated tasks, (D) and (E) for the Push-T task, and evaluate under setup (B) for real-world tasks. See Appendix~\ref{appx:hetero} for additional details.}
\label{fig:hetero}
    \end{center}
\vspace{-18pt}
\end{figure*}

\subsection{Baselines}
We follow the training and evaluation procedures described in Sections~\ref{s:train} and ~\ref{s:eval}. Baselines are categorized into two groups: representation learning methods for pre-training the vision encoder, and policy classes for fine-tuning with different policy head architectures and BC objectives. We include results for different combinations of the two groups and report them in Table \ref{tab:main_result}.

\textbf{Representation Learning}

\begin{itemize}[nosep,leftmargin=1em,labelwidth=*,align=left]
\item \textbf{Random, ImageNet, R3M}: ResNet-18 trained from scratch, or initialized with ImageNet-1K \cite{deng2009imagenet} or R3M \citep{nair2022r3m} pre-trained weights.
\item \textbf{TCN}: A single-view time-contrastive network \citep{sermanet2018time}.
\item \textbf{VINN}: A Bootstrap Your Own Latent (BYOL) \citep{grill2020bootstrap} pre-training on demonstrations \citep{pari2021surprising}.  
\item \textbf{DynaMo}: In-domain pre-training with inverse and forward dynamics model \citep{cui2024dynamo}.
\item \textbf{EquiVar}: Rotation group-equivariant encoder that improves data efficiency for DP \citep{wang2024equivariant}. % The encoder is modified to maintain equivariance with 224×224 resolution.
\end{itemize}

\textbf {Policy Class} When fine-tuning with the BC objective, we consider two policy classes: 
\begin{itemize}[nosep,leftmargin=1em,labelwidth=*,align=left]
\item \textbf{MLP}: A CNN-based policy head with a direct supervised learning objective. \item \textbf{DP} (Diffusion Policy): A CNN-based policy head trained with diffusion objective \citep{chi2024diffusionpolicy}. 
\end{itemize}

\subsection{Heterogeneous Data Collection Setup} 

Behavior cloning methods are commonly trained on demonstrations collected from a consistent viewpoint, often using a static, fixed camera—a condition that humans do not require. While this assumption simplifies the learning problem by enforcing visual consistency, it breaks down at scale, where data naturally spans diverse camera setups and collection conditions \citep{khazatsky2024droid}. Even minor variations in camera pose between demonstrations have been shown to substantially degrade performance \citep{xie2024decomposing, saxena2025matters}. To evaluate generalization under such visual shifts, we construct settings where each demonstration begins from a randomized camera pose, with optional movements during the episode to introduce additional viewpoint variation. For the Push-T task, we randomize object and background colors to test robustness to appearance shifts, as shown in Fig.~\ref{fig:hetero}. To evaluate compositional generalization across demonstrations, we do not apply rotation or color augmentations during training. For additional details, see Appendix \ref{appx:hetero}.

\begin{table}[t]
    \begin{center}
        \includegraphics[width=0.78\linewidth]{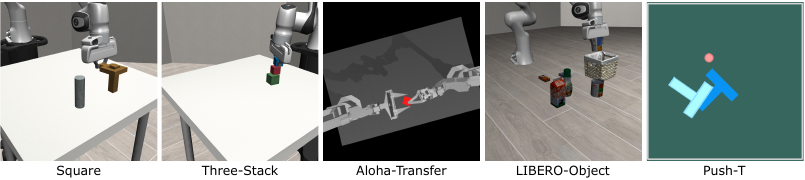}
        \label{fig:task}
    \end{center}
\centering
\renewcommand{\arraystretch}{1.2}
\resizebox{\textwidth}{!}{%
\begin{tabular}{l|cc|cc|cc|cc|cc}
\toprule
\multicolumn{1}{c}{} 
& \multicolumn{2}{c}{\textbf{Square}} 
& \multicolumn{2}{c}{\textbf{Three-Stack}} 
& \multicolumn{2}{c}{\textbf{Aloha-Transfer}} 
& \multicolumn{2}{c}{\textbf{LIBERO-Object}} 
& \multicolumn{2}{c}{\textbf{Push-T}} \\
\cmidrule(lr){2-3} \cmidrule(lr){4-5} \cmidrule(lr){6-7} \cmidrule(lr){8-9} \cmidrule(lr){10-11}

& Fixed-Cam & Dyn-Cam & Fixed-Cam & Dyn-Cam & Fixed-Cam & Dyn-Cam & Fixed-Cam & Dyn-Cam & Fixed-Color & Rand-Color \\
\bottomrule
\textbf{Rep-Only} & & & & & & & & & & \\
ImageNet & 0.20 & 0.00 & 0.00 & 0.00 & 0.26 & 0.02 & 0.82 & 0.59 & 0.00 & 0.00 \\
R3M   & 0.04 & 0.00 & 0.02 & 0.00 & 0.28 & 0.00 & 0.57 & 0.56 & 0.00 & 0.00 \\
TCN      & 0.53 & 0.07 & 0.28 & 0.01 & 0.17 & 0.0 & 0.90 & 0.92 & 0.04 & 0.02 \\
VINN     & 0.13 & 0.04 & 0.00 & 0.00 & 0.13 & 0.02 & 0.92 & 0.62 & 0.00 & 0.01 \\
DynaMo   & 0.01 & 0.00 & 0.00 & 0.00 & 0.11 & 0.09 & 0.58 & 0.57 & 0.00 & 0.02 \\
CLASS    & \textbf{0.93} & \textbf{0.70} & \textbf{0.94} & \textbf{0.77} & \textbf{0.91} & \textbf{0.56} & \textbf{0.99} & \textbf{0.99} & \textbf{0.85} & \textbf{0.66} \\
\midrule
\textbf{MLP} & & & & & & & & & & \\
ImageNet & 0.61 / 0.91 & 0.11 / 0.18 & 0.31 / 0.91 & 0.05 / 0.36 & 0.25 / \textbf{1.00} & 0.10 / 0.33 & 0.99 / \textbf{1.00} & 0.60 / 0.46 & 0.66 / 0.87 & 0.21 / 0.27 \\
CLASS    & \textbf{0.88} / \textbf{0.95} & \textbf{0.67} / \textbf{0.62} & \textbf{0.62} / \textbf{0.93} & \textbf{0.61} / \textbf{0.80} & \textbf{0.72} / \textbf{1.00} & \textbf{0.52} / \textbf{0.78} & \textbf{1.00} / 0.99 & \textbf{1.00} / \textbf{0.96} & \textbf{0.82} / \textbf{0.89} & \textbf{0.69} / \textbf{0.64} \\
\midrule
\textbf{DP} & & & & & & & & & & \\
Random   & 0.45 / 0.85 & 0.06 / 0.06 & 0.53 / 0.86 & 0.05 / 0.28 & 0.32 / \textbf{1.00} & 0.11 / 0.15 & 0.99 / 0.91 & 0.57 / 0.04 & 0.62 / 0.89 & 0.04 / 0.16 \\
ImageNet & 0.72 / 0.91 & 0.09 / 0.21 & 0.86 / 0.94 & 0.07 / 0.61 & 0.37 / \textbf{1.00} & 0.09 / 0.20 & \textbf{1.00} / 0.97 & 0.64 / 0.19 & 0.70 / 0.92 & 0.15 / 0.53 \\
R3M      & 0.42 / 0.88 & 0.04 / 0.14 & 0.36 / 0.84 & 0.02 / 0.36 & 0.35 / \textbf{1.00} & 0.10 / 0.09 & 0.99 / 0.94 & 0.60 / 0.12 & 0.65 / 0.89 & 0.00 / 0.05 \\
TCN      & 0.78 / 0.91 & 0.17 / 0.34 & 0.85 / 0.93 & 0.12 / 0.71 & 0.52 / \textbf{1.00} & 0.11 / 0.33 & 0.98 / \textbf{0.99} & 0.95 / 0.96 & 0.69 / 0.96 & 0.33 / 0.49 \\
VINN     & 0.57 / 0.87 & 0.07 / 0.07 & 0.63 / 0.88 & 0.01 / 0.22 & 0.39 / \textbf{1.00} & 0.08 / 0.15 & 0.92 / \textbf{0.99} & 0.57 / 0.06 & 0.66 / 0.93 & 0.13 / 0.06 \\
DynaMo   & 0.39 / 0.80 & 0.07 / 0.05 & 0.48 / 0.81 & 0.02 / 0.28 & 0.82 / 0.88 & 0.17 / 0.23 & 0.96 / \textbf{0.99} & 0.46 / 0.40 & 0.67 / 0.90 & 0.06 / 0.24 \\
EquiVar  & 0.77 / 0.94 & 0.03 / 0.08 & 0.43 / 0.95 & 0.06 / 0.36 & -- & -- & 0.99 / \textbf{0.99} & 0.61 / 0.10 & -- & -- \\
CLASS    & \textbf{0.91} / \textbf{0.95} & \textbf{0.64} / \textbf{0.68} & \textbf{0.89} / \textbf{0.97} & \textbf{0.76} / \textbf{0.93} & \textbf{0.83} / 0.99 & \textbf{0.61} / \textbf{0.95} & \textbf{1.00} / \textbf{0.99} & \textbf{1.00} / \textbf{0.97} & \textbf{0.81} / \textbf{0.96} & \textbf{0.69} / \textbf{0.70} \\
\bottomrule
\end{tabular}
}
\vspace{2pt}
\caption{\textbf{Performance comparison between CLASS and baselines on simulation tasks.} We report the maximum task success rate across 50 randomly initialized scenes except for LIBERO-object which tests each task 20 times, averaged over three seeds. All methods are evaluated ten times throughout training at regular intervals. Each cell shows \textbf{non-parametric / parametric} evaluation when applicable. For Rep-Only settings, only non-parametric results are reported. All results use a single global-view camera. Whenever applicable, ImageNet pre-training weights are used for representation learning. For additional discussions and evaluations, see Appendix \ref{appx:add_exp}.}
\label{tab:main_result}
\vspace{-18pt}
\end{table}

\begin{figure*}[b]
    \begin{center}
        \includegraphics[width=0.99\linewidth]{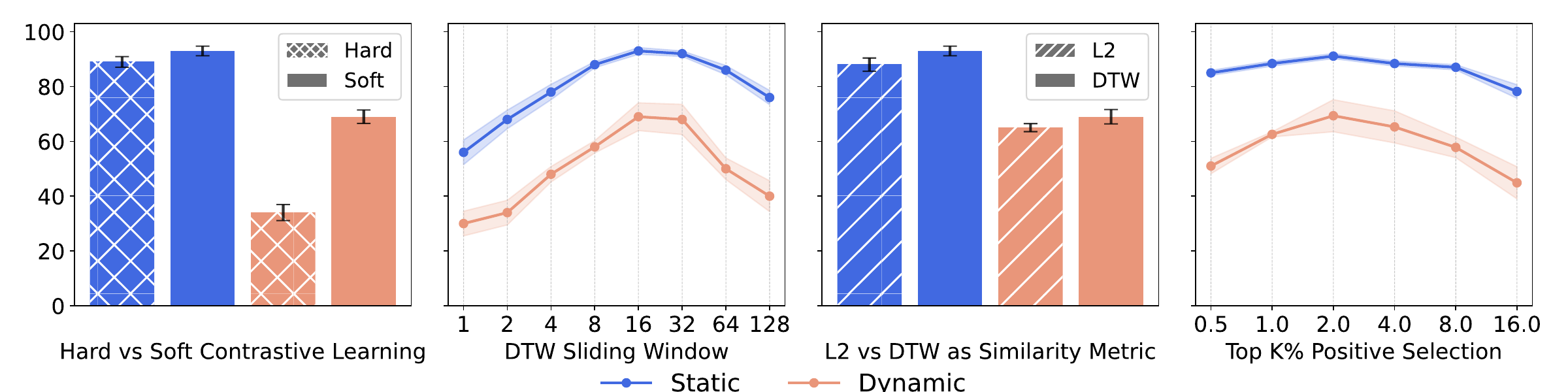}
\caption{\textbf{Design Choice Analysis}: Success rate of Fixed-Cam (blue) and Dynamic-Cam (orange) for Square task with Rep-Only as a function of (a) hard or soft contrastive learning, (b) size of DTW sliding window, (c) similarity metric, and (d) the quantile K (\%) value for positive sample selection.}
        \label{fig:analysis}
    \end{center}
\end{figure*}

\subsection{Simulation Results}
Table~\ref{tab:main_result} summarizes the simulation performance of CLASS compared to baseline methods. CLASS consistently improves success rates across all tasks and policy classes: Across the static and dynamic settings, it achieves \textbf{85\%} success rate with \textbf{MLP} and \textbf{91\%} with \textbf{Diffusion Policy (DP)} on average, significantly outperforming the best baselines, which reach only 63\% and 77\%, respectively. 
Larger gains are observed under dynamic camera and random object color setups, where CLASS achieves mean success rates of \(76\%\) (MLP) and \(85\%\) (DP), compared to \(32\%\) and \(57\%\) from the best baselines.
In the non-parametric (Rep-Only) setting, CLASS achieves success rates comparable to parametric DP, despite using no policy head, with only a \(9\%\) drop in mean success rate with an average of \(83\%\) across all tasks.
Finally, strong performance on Aloha-Transfer and LIBERO-Object demonstrates that CLASS generalizes effectively to different action spaces and multi-task learning scenarios.
% For non-parametric evaluation, we select \(\tau_{nn}\) from \{0.01, 0.02, 0.04\} based on the value that achieves the lowest retrieval error on the training dataset.

\subsection{Design Choice Analysis for CLASS}
\label{sec:ablation}

We evaluate the impact of key design choices in CLASS, including soft contrastive learning, the use of a sequence-level similarity metric, the choice of DTW for measuring action similarity, and the positive pair selection threshold. 
Fig.~\ref{fig:analysis}(a) compares soft contrastive learning against hard contrastive learning, where the soft positive weights \(w_{ij} > 0\) in Equation~\ref{eq:soft_info_nce} are replaced with uniform weights of 1. Critically, it is shown that removing the graded similarity signals leads to significant degradation of performance. 
Fig.~\ref{fig:analysis}(b) shows that increasing the sliding window size for calculating DTW between action sequences improves success rates up to \(T = 16\), indicating the importance of using sequence-level similarity. 
As shown in Fig.~\ref{fig:analysis}(c), DTW outperforms L2 distance in both static and dynamic settings. 
Finally, varying the positive selection quantile in Fig.~\ref{fig:analysis}(d) reveals a tradeoff: small thresholds limit available information, while large thresholds lead to more false positives. See Appendix~\ref{appx:implementation} for implementation details.

\subsection{Real-World Experiment / Results}
\begin{wrapfigure}[9]{r}{0.37\linewidth}
    \centering
    \vspace{-49pt}
    \includegraphics[width=\linewidth]{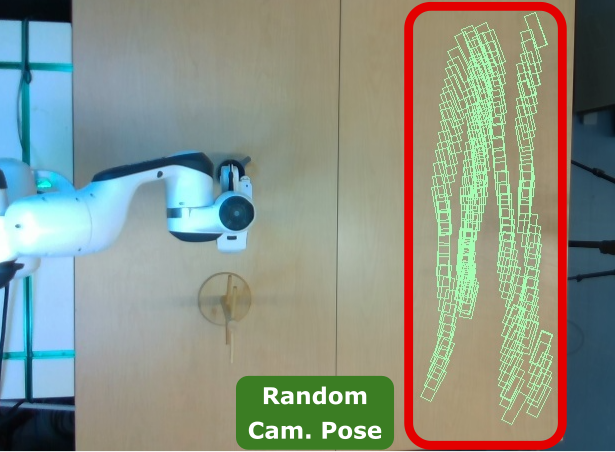} 
    \label{fig:camera}
    \vspace{-16pt}
\caption{Visualization of camera poses for the Mug-Hang task.}
    \label{fig:camera}
\end{wrapfigure}

We design three real-world manipulation tasks as described in Section~\ref{ss:tasks}, using a Franka Research 3 robot and a single RealSense D435 camera capturing 720p RGB images at 30 FPS. The images are cropped and resized to \(256 \times 256 \times 3\). A proficient expert teleoperates the robot using a VR controller. To introduce visual variation, the camera is mounted on a tripod and randomly repositioned after each demonstration, as illustrated in Fig.~\ref{fig:camera}. For evaluation, we compare Diffusion Policy pre-trained on ImageNet (ImageNet-DP) with CLASS-DP, which uses representations learned via the CLASS objective. All results use a single global-view camera.

\begin{figure}[t]
    \begin{center}
    \includegraphics[width=\linewidth]{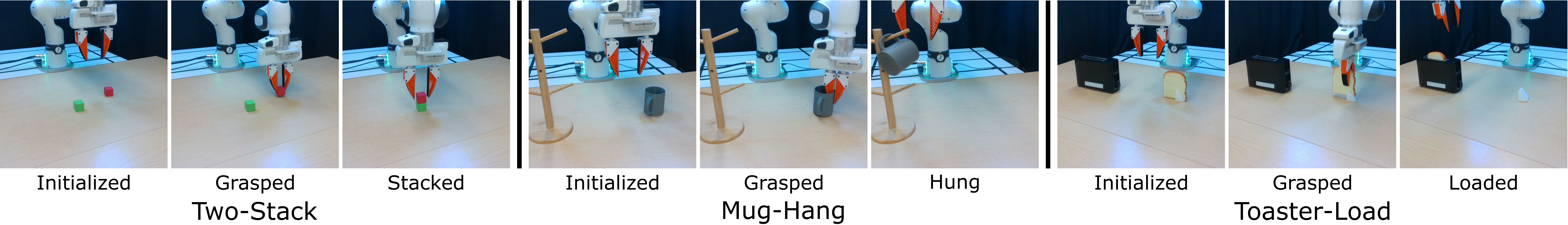}
    \label{fig:task}
    \vspace{-16pt}
    \caption{\textbf{Real-world task execution.} The initial scenes and the subtasks are visualized.}
    \end{center}
    \vspace{-10pt}
\end{figure}

\begin{table}[t]
  \centering
  \resizebox{0.8\linewidth}{!}{%
    \begin{tabular}{lcccccc}
      \toprule
      & \multicolumn{2}{c}{\textbf{Two-Stack} (real)} 
      & \multicolumn{2}{c}{\textbf{Mug-Hang} (real)} 
      & \multicolumn{2}{c}{\textbf{Toaster-Load} (real)} \\
      \cmidrule(lr){2-3} \cmidrule(lr){4-5} \cmidrule(lr){6-7}
      Method 
        & Grasped & Stacked
        & Grasped & Hung 
        & Grasped & Loaded \\
      \midrule
      ImageNet-DP    
        &  0.15 / 0.30   &  0.00 / 0.10    
        &  0.15 / 0.05   &  0.00 / 0.00 
        &  0.95 / \textbf{1.00}   &  0.00 / 0.05    \\
      CLASS-DP
        & \textbf{0.70 / 0.80} & \textbf{0.45 / 0.60} 
        & \textbf{0.75 / 0.80} & \textbf{0.55 / 0.65} 
        & \textbf{1.00 / 1.00} & \textbf{0.35 / 0.55} \\
      \bottomrule
    \end{tabular}%
  }
  \vspace{6pt}
  \caption{The performance of CLASS pre-trained DP for real-world tasks in comparison to ImageNet pre-trained DP. We report the success rate for each subtask over 20 randomly initialized scenes and camera placements. Each cell shows the results using the non-parametric / parametric evaluations.}
  \label{tab:real_tasks}
  \vspace{-14pt}
\end{table}

Table~\ref{tab:real_tasks} summarizes performance on real-world tasks. CLASS-DP outperforms ImageNet-DP across all three tasks, achieving on average \textbf{37\%} and \textbf{45\%} higher subtask success and final task completion rates, respectively, in the non-parametric setting. In the parametric setting, the differences increase to \textbf{41\%} and \textbf{55\%}, respectively. These results demonstrate that CLASS generalizes effectively to real-world settings and improves policy performance when learning from demonstrations with varying viewpoints.

%===============================================================================

\section{Conclusion}
\label{sec:conclusion}
This paper presents CLASS, a novel framework for imitation learning in robotic manipulation based on soft contrastive learning with an action sequence similarity metric. We confirm that BC suffers from its strong end-to-end supervision mechanism, especially when learning from heterogeneous datasets. In contrast, CLASS learns representations that map observations with similar trajectory-level behaviors closer in feature space, enabling robust representation learning in both homogeneous and heterogeneous settings. Across 5 simulated and 3 real-world tasks, CLASS consistently outperforms baselines in both retrieval-based evaluations and downstream policy fine-tuning.
Future directions include extending CLASS to large-scale policy learning and cross-embodiment learning with a unified action space, leveraging diverse offline data to enable few-shot learning, and investigating its effectiveness when learning from suboptimal or noisy demonstrations.

\clearpage
\section*{Limitations}
While CLASS demonstrates strong performance across a range of tasks and settings, several limitations remain. 
First, CLASS requires additional pre-computing pairwise DTW distances to define positive pairs, which has a quadratic cost in the number of data points. However, CLASS's faster convergence during training can make it more computationally efficient in repeated training runs. For more detailed analysis, see Appendix \ref{appx:train_speed}.
Second, the use of suboptimal or noisy demonstrations is not addressed in this work.
Lastly, the current scope of the work is limited to vision modality.
% Lastly, CLASS tends to underperform in low-data regimes. When only a few demonstrations are available, the quality and density of positive pairs degrade, limiting the strength of the learned representations. 

\textbf{Failure modes} 
On real-world tasks, the most common failure mode involves the robot proceeding to the next subtask region without completing the current one. For example, in the Mug-Hang task, the policy occasionally fails to grasp the mug but still moves toward the hanger with the gripper closed. During non-parametric evaluations, we observe less fine-grained control, particularly on high-precision tasks such as Toaster-Load, where it fails to accurately insert the bread into the slot. Additionally, when the camera is positioned outside the pre-defined region used during data collection, performance degrades significantly, indicating that the policy still has a limited extrapolative generalization capability to unseen viewpoints.

\section*{Acknowledgment}
This research was partly supported by Delta Electronics Inc., Toyota Research Institute, and NSF CMMI-2443076.
We acknowledge Research Computing at the University of Virginia for providing the computational resources that made the results in this work possible.

%===============================================================================

% \[
% \log p_{ij} = S_{ij} - \log \sum_{k=1, k \ne i}^{B} \exp{S_{ik}}.
% \]
% \begin{equation}
% \mathcal{L}_{\text{Soft InfoNCE}} = -\frac{1}{B}\sum_{i=1}^{B} 
% \frac{\sum_{j=1}^{B} w_{ij}\log p_{ij}}{\sum_{j=1}^{B} w_{ij}}.
% \label{eq:soft_info_nce}
% \end{equation}

\clearpage
\bibliography{bibliography} 
\clearpage
% The acknowledgments are automatically included only in the final and preprint versions of the paper.
% \acknowledgments{If a paper is accepted, the final camera-ready version will (and probably should) include acknowledgments. All acknowledgments go at the end of the paper, including thanks to reviewers who gave useful comments, to colleagues who contributed to the ideas, and to funding agencies and corporate sponsors that provided financial support.}
%===============================================================================
\appendix

\section{Proof of Proposition 1}
\label{appx:proposition}
%\textbf{Proof.} 
Let $q_{ij} = w_{ij} / Z$ where $Z = \sum_{j=1}^{B} w_{ij}$ is the partition function.
We can write the instance-wise proposed loss as follows:
\begin{equation*}
    l_{\text{CL}}^{(i)} = -
\frac{\sum_{j=1}^{B} w_{ij}\log p_{ij}}{\sum_{j=1}^{B} w_{ij}}
    = -\sum_{j=1}^{B} \frac{w_{ij}}{Z}\log p_{ij}
    = \sum_{j=1}^{B} q_{ij} \log \frac{q_{ij}}{p_{ij}} - \underbrace{q_{ij}\log q_{ij}}_{\text{constant}}
\end{equation*}
Let $P$ and $Q$ be the probability distribution of $p_{ij}$ and $q_{ij}$.
We can rewrite the instance-wise loss as
\begin{equation*}
    l_{\text{CL}}^{(i)} = D_{\text{KL}}(Q||P) + \text{const}
\end{equation*}
This is the KL divergence of the predicted softmax similarity from the soft positive assignments.

\section{DTW-based Similarity Calculation}
\label{appx:precompute}
To compute DTW between action sequences, we use \texttt{aeon}~\citep{aeon24jmlr}, a Python toolkit that supports efficient DTW computation for multivariate time-series data. For the Aloha-Transfer environment, each action sequence is 8-dimensional, comprising 7D joint positions and a 1D gripper action. In Push-T, actions are 2-dimensional, controlling the absolute 2D target position. For the other environments that use end-effector pose control, each action sequence is 10-dimensional, consisting of 3D position, 6D orientation, and 1D gripper action. To balance the influence of position and orientation, orientation values are scaled by a factor of 0.5 prior to the DTW calculations.

To enable consistent interpretation of DTW distances, we modeled the empirical cumulative distribution function (CDF) from a set of pairwise DTW values. In our experiments, we precomputed these distances for all anchor-positive pairs. As noted in the main text, this approach is more efficient than constructing a full \( N \times N\) distance matrix and was feasible for the datasets used. 

However, we recognize that this pre-computation method is not fully scalable. While it avoids the full distance matrix, the computational and memory cost still scales quadratically with the number of anchor-positive pairs, becoming prohibitive for large datasets. For these scenarios, an online DTW computation strategy offers a potential alternative. This method would estimate the CDF from a data sample and compute pairwise distances on-the-fly within training batches. This bypasses the need for pre-computation and large-scale storage, as the computational overhead is tied to the much smaller batch size, thereby extending the scalability of our method to larger datasets.

\section{Implementation Details and Hyperparameters}
\label{appx:implementation}
% \subsection{Contrastive Learning via Action Sequence Supervision (CLASS)}
% \label{appx:class}

\subsection{Representation Learning}
\label{appx:rep_learning}
We focus on training the vision encoder for representation learning. Except for Equivariant Diffusion Policy, all methods use a ResNet-18 backbone followed by a spatial softmax layer. All batch normalization layers are replaced with group normalization layers. 

\begin{table}[h]
\centering
\begin{tabular}{l c}
\toprule
\textbf{Name} & \textbf{Value} \\
\midrule
% Optimizer & AdamW \citep{loshchilov2017decoupled} \\
% Learning rate & $1e^{-4}$ \\
% Weight decay & 0.01 \\
% Betas & (0.9, 0.99) \\
Gradient clip norm & 0.5 \\
Scheduler & Cosine \\
Warmup steps & 500 \\
Use EMA & True \\ 
EMA Power & 0.75 \\
\bottomrule
\end{tabular}
\vspace{5pt}
\caption{Shared hyperparameters for representation learning methods.}
\end{table}

\textbf{Contrastive Learning via Action Sequence Supervision (CLASS)} \\
Table~\ref{tab:class_shared} lists the shared training and evaluation hyperparameters used across all tasks. 
% The temperature $\tau_{nn}$ for retrieval evaluation is selected from \{0.01, 0.02\} based on which yields the lowest validation retrieval error. 
$T_{a}$ indicates the rollout horizon for action sequences during evaluation.  Table~\ref{tab:class_task_specific} shows task-specific settings for Top K (\%) similarity thresholds for positive sample selection and number of pre-training epochs. Epochs are tuned based on dataset size and validation performance. 

\begin{table}[h]
\centering
\begin{tabular}{l c}
\toprule
\textbf{Name} & \textbf{Value} \\
\midrule
\textbf{Pre-computing} &  \\
DTW Horizon & 16 \\
 &  \\
\textbf{Training} &  \\
% Optimizer & SGD \cite {robbins1951stochastic} \\
Optimizer & LARS \cite {you2017large} \\
Batch size & 160 \\
Learning rate & 0.4 \\
Momentum & 0.9 \\
Weight decay & $1e^{-6}$ \\
% Projection dim & 1024 \\
$\tau$ & 0.05 \\
% $\rho$ & 0.01 \\
 &  \\
\textbf{Non-parametric Evaluation} &  \\
$T_{a}$ & 16 \\
$k_{nn}$ & 64 \\

\bottomrule
\end{tabular}
\vspace{5pt}
\caption{Shared hyperparameters for CLASS.}
\label{tab:class_shared}
\end{table}

\begin{table}[h]
\centering
\begin{tabular}{lccc}
\toprule
\textbf{Task} & \textbf{Top K (\%)} & \textbf{Epochs} & $\mathbf{\tau}_{nn}$   \\

\midrule
Square & 2.5 & 50 & 0.01 / 0.02 \\
Three-Stack & 1.0 & 10 & 0.01 / 0.02 \\
Aloha-Transfer & 2.5 & 50 & 0.01 / 0.02 \\
LIBERO-Object & 2.5 & 50 & 0.01 / 0.02 \\
Push-T & 1.5 & 100 & 0.001 / 0.002 \\
Two-Stack & 2.5 & 50 & 0.02 \\
Mug-Hang & 2.5 & 50 & 0.02 \\
Toaster-Load & 2.5 & 50 & 0.02 \\
\bottomrule
\end{tabular}
\vspace{5pt}
\caption{Task-specific hyperparameters for CLASS. Homogeneous and heterogeneous settings are distinguished in the  $\mathbf{\tau}_{nn}$ values in each cell.
}
\label{tab:class_task_specific}
\end{table}

\textbf{Time Contrastive Network (TCN)} \\
We use the single-view TCN implementation from \href{https://github.com/kekeblom/tcn}{https://github.com/kekeblom/tcn}. The model is trained for 200 epochs with a batch size of 64 and a learning rate of 1$e^{-4}$, where each batch consists of 64 triplets of images used for contrastive learning. %In the Dynamic Camera setting, performance tend to degrade with additional training and the better results are usually attained with smaller epochs.

% \subsubsection{Masked AutoEncoder (MAE)}

% We follow the official implementation from \href{https://github.com/facebookresearch/mae}{https://github.com/facebookresearch/mae} with a mask ratio of 75\%. The model is trained for 300 epochs with a batch size of 64 by reconstructing masked patches from visible ones. For feature extraction, the [CLS] token from the final encoder layer is used as the image representation.

\textbf{Visual Imitation through Nearest Neighbors (VINN)} \\
We follow the BYOL implementation from \href{https://github.com/lucidrains/byol-pytorch} {https://github.com/lucidrains/byol-pytorch}. The model is trained for 200 epochs with a batch size of 64 and a learning rate of 1$e^{-4}$.

\textbf{Dynamics pre-training for Visuo-Motor Control (DynaMo)} \\
We follow the procedures from the original paper.
The encoder is based on a ResNet-18 backbone and is trained for 40 epochs with a batch size of 64 and a learning rate of 1$e^{-4}$.

\begin{table}[h]
\centering
\begin{tabular}{l c}
\toprule
\textbf{Name} & \textbf{Value} \\
\midrule
Epochs & 40 \\
Batch size & 64  \\
% Learning rate & $1e^{-4}$ \\
Obs. context & 5 \\
Forward dynamics dropout & 0.0 \\
Transition latent dim & 32 \\

\bottomrule
\end{tabular}
\vspace{5pt}
\caption{Hyperparameters for DynaMo.}
\end{table}

\subsection{Policy Learning}

% All methods use a two-stage pipeline: encoder pre-training followed by policy fine-tuning. For homogeneous settings, fine-tuning the encoder with the same learning rate as for the policy head along with the policy head consistently yields better performance across baselines, and we report those results in Figure~\ref{tab:main_result}. For heterogeneous settings, w
For all experiments, we choose reduced the learning rate for the vision encoder to better preserve the pre-trained representation. The learning rate for the vision encoder is {$1e^{-5}$} and the learning rate for the policy head is {$1e^{-4}$}. 

% \begin{table}[h]
% \centering
% \renewcommand{\arraystretch}{1.5}
% \resizebox{0.9\textwidth}{!}{
% \begin{tabular}{|>{\centering\arraybackslash}p{3cm}|c|c|c|c|}
% \hline
% \textbf{Component} & \textbf{Fixed Camera} & \textbf{Dynamic Camera} & \textbf{Fixed Color} & \textbf{Random Color} \\
% \hline
% Vision Encoder
%   & \multirow{2}{*}{$1e^{-4}$} 
%   & $1e^{-5}$ 
%   & \multirow{2}{*}{$1e^{-4}$} 
%   & $1e^{-5}$  \\
% \cline{3-3}
% \cline{1-1}
% \cline{5-5}
% Policy Head
%   & 
%   & $1e^{-4}$ 
%   & 
%   & $1e^{-4}$ 
%   \\
% \hline
% \end{tabular}
% }
% \vspace{5pt}
% \caption{Fine-tuning learning rates for vision encoder and policy head across settings.}
% \label{tab:lr}
% \end{table}

\textbf{MLP} \\
The architecture is based on a U-Net structure adapted from the CNN-based Diffusion Policy (DP), modified to support action chunking with direct action sequence supervision. Table \ref{tab:policy_shared} includes the policy hyperparameters.

\textbf{Diffusion Policy} \\
We adopt the CNN-based model architecture from \citet{chi2024diffusionpolicy}. Table \ref{tab:dp_hyperparam} includes diffusion-specific hyperparameters for regular DP and Equivariant DP.

\textbf{Equivariant Diffusion Policy (EquiVar)} \\
We focus on the equivariant vision encoder as a replacement of the ResNet-18 encoder, following the official implementation from \href{https://github.com/pointW/equidiff/}{https://github.com/pointW/equidiff}. The vision encoder is trained from scratch for policy training. The model architecture is modified to accept input images of size $224 \times 224 \times 3$.

\begin{table}[h]
\centering
\begin{tabular}{l c}
\toprule
\textbf{Name} & \textbf{Value} \\
\midrule
\textbf{Training} &  \\
Batch size & 64 \\
Weight decay & $1e^{-6}$ \\
Betas & (0.9, 0.99) \\
Gradient clip norm & 0.5 \\
Scheduler & Cosine \\
Warmup steps & 500 \\
Use EMA & True \\ 
EMA Power & 0.75 \\
$T_{o}$ & 1 \\
$T_{p}$ & 16 \\
 &  \\
\textbf{Parametric Evaluation} &  \\
$T_{a}$ & 12 \\
\bottomrule
\end{tabular}
\vspace{5pt}
\caption{Shared policy hyperparameters for MLP and DP.}
\label{tab:policy_shared}
\end{table}

% \begin{table}[h]
% \centering
% \caption{MLP-specific hyperparameters}
% \begin{tabular}{l c}
% \toprule
% \textbf{Name} & \textbf{Value} \\
% \midrule
% \textbf{Training} &  \\
% Epochs & 200 \\
% \bottomrule
% \end{tabular}
% \label{tab:mlp_hyperparam}
% \end{table}

\begin{table}[h]
\centering
\begin{tabular}{l c}
\toprule
\textbf{Name} & \textbf{Value} \\
\midrule
\textbf{Training} &  \\
% Epochs & 500 \\
Model Type & DDIM \cite {song2020denoising}\\
Prediction Type & epsilon \\
Denoising Step & 16 \\
&  \\
\textbf{Parametric Evaluation} &  \\
Denoising Step & 16 \\
\bottomrule
\end{tabular}
\vspace{5pt}
\caption{Diffusion-specific hyperparameters.}
\label{tab:dp_hyperparam}
\end{table}

\begin{table}[h]
\centering
\begin{tabular}{lcc}
\toprule
\textbf{Task} & \textbf{Training Epochs} & \textbf{Fine-tuning Epochs} \\

\midrule
Square & 500 & 100 \\
Three-Stack & 100 & 20 \\
Aloha-Transfer & 500 & 400 \\
LIBERO-Object & 500 & 100 \\
Push-T & 500 & 250 \\
Two-Stack & 500 & 100 \\
Mug-Hang & 500 & 100 \\
Toaster-Load & 500 & 100 \\
\bottomrule
\end{tabular}
\vspace{3pt}
\caption{Task-specific hyperparameters for policy head training. Training Epochs denote the number of epochs used to train the BC policy head for the baseline methods. Fine-tuning Epochs indicate the number of epochs used to fine-tune the policy head after CLASS pre-training. CLASS pre-training enables faster convergence speed during fine-tuning and thus requires less number of epochs. 
}
\label{tab:bc_task_specific}
\end{table}

\section{Environments Details}
\label{appx:env}
All environments use position-based control. For tasks involving control of the end-effector pose, orientation is represented using the continuous 6D representation proposed by \citet{zhou2019continuity}. %All images are resized to \(256 \times 256 \times 3\).

\begin{table}[h]
\centering
\label{tab:method_config}
\setlength{\tabcolsep}{1pt}  % Tighter column spacing
\resizebox{\textwidth}{!}{%
\begin{tabular}{lcccc}
\toprule
\textbf{Task Name} & \textbf{Observation Modalities} & \textbf{Image Resolution} & \textbf{Action Space}  & \textbf{Max Steps}   \\
\midrule
\textbf{Simulation} \\
Square & Agent-View Image, End-Effector Pose, Joint Position, Gripper State &\(256 \times 256 \times 3\) & End-Effector Position  & 400 \\
Three-Stack & Agent-View Image, End-Effector Pose, Joint Position, Gripper State &\(256 \times 256 \times 3\)  & End-Effector Position & 600 \\
Aloha-Transfer & Top-Down Image, Joint Position, Gripper State &\(256 \times 256 \times 3\)  & Bi-manual Joint Position  & 600  \\ 
Libero-Object & Agent-View Image, End-Effector Pose, Joint Position, Gripper State &\(256 \times 256 \times 3\)  & End-Effector Position  & 400 \\
Push-T & Top-Down Image, Agent Position &\(256 \times 256 \times 3\)  & 2D Position  & 500  \\
\midrule
\textbf{Real-world} \\
Two-Stack & Agent-View Image, End-Effector Pose, Joint Position, Gripper State &\(256 \times 256 \times 3\) & End-Effector Position  & 600 \\
Mug-Hang & Agent-View Image, End-Effector Pose, Joint Position, Gripper State &\(256 \times 256 \times 3\) & End-Effector Position  & 600 \\
Toaster-Load & Agent-View Image, End-Effector Pose, Joint Position, Gripper State &\(256 \times 256 \times 3\) & End-Effector Position  & 600 \\
\bottomrule
\end{tabular}
}
\vspace{3pt}
\caption{Observation modalities, image resolution, action space, and maximum episode length for each task.}
\end{table}

\section{Heterogeneous Data Illustration}
\label{appx:hetero}
\subsection{Simulation Camera Placement}
For LIBERO-Object, Square, and Three-Stack tasks, we implement a moving camera by attaching it to a mocap body and updating its pose at every environment step. The camera is initialized at a random position on a sphere of radius 1.0 centered at a fixed look-at target. The azimuth \(\theta\) is sampled from \([0, 2\pi)\), and the elevation \(\phi\) from \([\pi/3, 2\pi/3]\), ensuring relatively horizontal views. The orientation is set to point at the target using a look-at transform with world up-vector \([0, 0, 1]\).
At the start of each episode, a direction vector is sampled from \(x \in [-1, 1]\), \(y \in [-0.01, 0.3]\), then projected onto the tangent plane at the initial camera position. The camera moves along this direction by rotating the position vector around the perpendicular axis, with a fixed arc step of $1.5e^{-4}$ per timestep. The orientation is updated each step to remain focused on the target. 

For the \textsc{Aloha-Transfer} task, we simulate camera motion by introducing a consistent per-episode image rotation on the top-down view. At the start of each episode, an initial rotation angle is uniformly sampled from \([0^\circ, 360^\circ)\), and a fixed rotation direction of either \(+0.5^\circ\) or \(-0.5^\circ\) per timestep is chosen. At each step, the image is rotated accordingly. 

\subsection{Object/Background Color}

For the \textsc{Push-T} task, we introduce per-episode color variation by randomly sampling RGB values for the background, goal region, tee block, and agent. The background color is constrained to avoid strong red tones. To ensure visibility, the goal, tee, and agent colors are sampled to be visually distinct from the background and from each other, based on a minimum Euclidean distance in RGB space. Colors remain fixed throughout each episode but vary across episodes.
\vspace{10pt}
\subsection{Real-world Camera Placement}

\begin{figure*}[h]
    \begin{center}
        \includegraphics[width=1.0\linewidth]{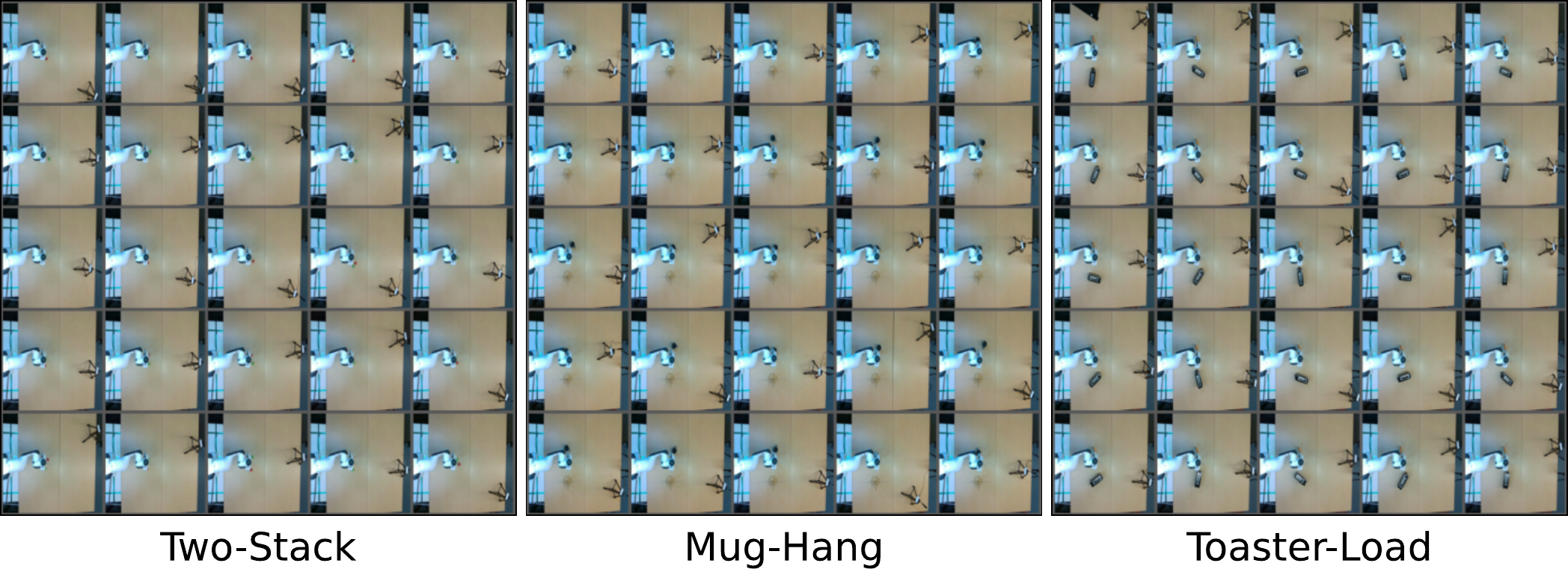}
\caption{
Top-down views from 25 randomly selected episodes, illustrating the distribution of real-world camera placements across the three real-world tasks.
}

        \label{fig:rw_camera}
    \end{center}
\end{figure*}

Figure~\ref{fig:rw_camera} shows top-down views from 25 randomly sampled episodes across the three real-world tasks. The variation in camera placement reflects the heterogeneity in data collection setups.

\section{Additional Experiments}
\vspace{10pt}

\label{appx:add_exp}

To evaluate the compositionality of the learned representation and to challenge methods to generalize without additional observations—such as a wrist-view camera—we constrain our original experimental setup to a single global view camera and apply image augmentations excluding color and rotational transformations. In this section, we demonstrate that incorporating a wrist-view camera or applying additional task-aligned data augmentations—both expected to benefit Behavior Cloning (BC)—also enhances CLASS performance. Crucially, CLASS still retains its advantages over BC under these conditions.

\vspace{14pt}
\subsection{Additional Wrist-View Camera}
\vspace{4pt}
In Figure \ref{fig:add_exp}(a), we evaluate adding a wrist-mounted camera for the Dynamic Camera Square task. The wrist camera provides a consistent viewpoint that is invariant to the global camera placement variations tested in our main study, which is expected to boost performance for behavior cloning.

With the wrist camera view added, the performance of fine-tuned policy with CLASS pre-training increases the from 68\% to 90\% success rate, while BC improves from 21\% to 66\%. This demonstrates that while providing a stable, ego-centric view does help both methods, CLASS still significantly outperforms BC.

\vspace{14pt}
\subsection{Additional Color Augmentation}
\vspace{4pt}
Figure \ref{fig:add_exp}(b) examines color augmentation on the Rand-Color Push-T task, where object colors are randomized to test visual robustness. We apply random hue, saturation, and brightness shifts during training, which is expected to improve generalization by reducing overfitting to specific color patterns.
With color augmentation, the fine-tuned policy with CLASS pre-training increases from 72\% to 90\% success rate, while BC improves from 53\% to 78\%. This confirms that while data augmentation benefits both approaches, CLASS pre-training continues to provide substantial performance gains even when visual robustness is enhanced through augmentation.

\begin{figure*}[h]
    \begin{center}
        \includegraphics[width=1.\linewidth]{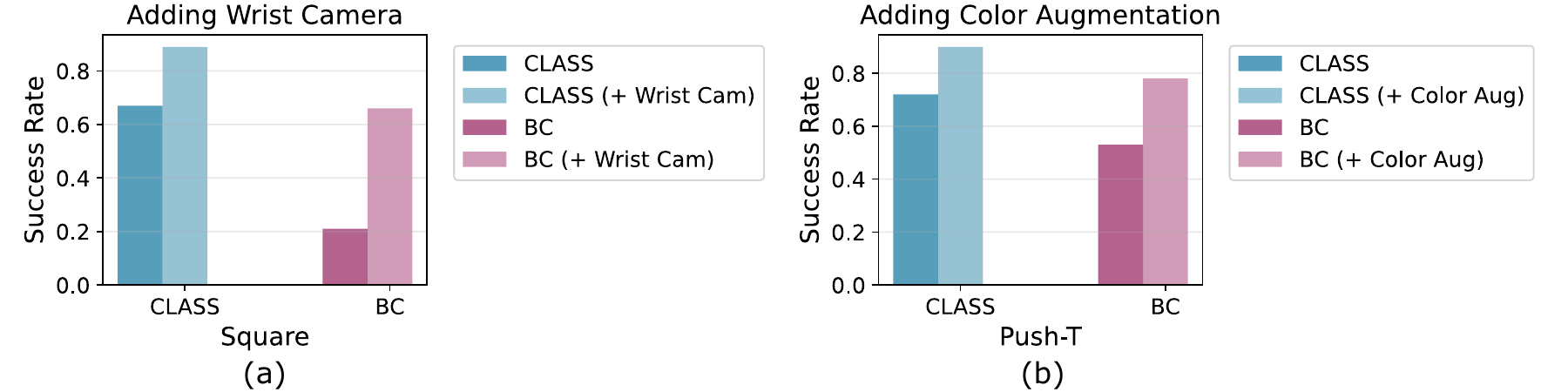}
\caption{Impact of adding wrist camera and task-aligned augmentations. (a) Success rates of the Dynamic Camera Square task with and without adding the wrist camera. (b) Success rates of the Random Color Push-T task with and without color augmentation during training. Both experiments are reported using parametric evaluation with BC fine-tuned policies with and without CLASS pre-training.}
        \label{fig:add_exp}
    \end{center}
    \vspace{-2pt}
\end{figure*}

\subsection{Benefits of ImageNet Pre-training for CLASS}
\vspace{4pt}
In our paper, all representation methods utilize ImageNet pre-training whenever applicable. In this section, we show the benefits of using ImageNet pre-trained weights for CLASS. We conduct an ablation study where we compare the policy performance using CLASS pre-training with and without ImageNet pre-trained Resnet-18 as initial weights for Square, Three-Stack, and Push-T tasks.
As shown in Figure~\ref{fig:add_exp}, training from scratch leads to significant drops in success rates across all three tasks. The performance drop is more pronounced in heterogeneous settings, including both Dynamic Camera and Random Color settings. Most significantly, training from scratch for Random Color Push-T task leads to 64\% drop in performance in comparison to using the ImageNet pre-trained weights.

\begin{figure*}[h]
    \begin{center}
        \includegraphics[width=0.7\linewidth]{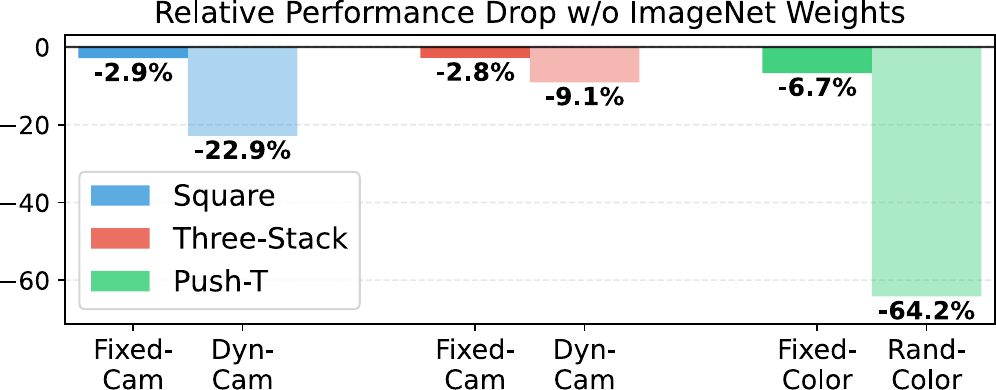}
\caption{The relative performance drop when training from scratch instead of using ImageNet pre-trained weights for CLASS pret-raining. The results are reported using non-parametric rollouts, averaged over three seeds.}
        \label{fig:add_exp}
    \end{center}
    \vspace{-2pt}
\end{figure*}

\section{Training Dynamics}
\label{appx:train_speed}

\begin{figure*}[h]
    \begin{center}
        \includegraphics[width=1.0\linewidth]{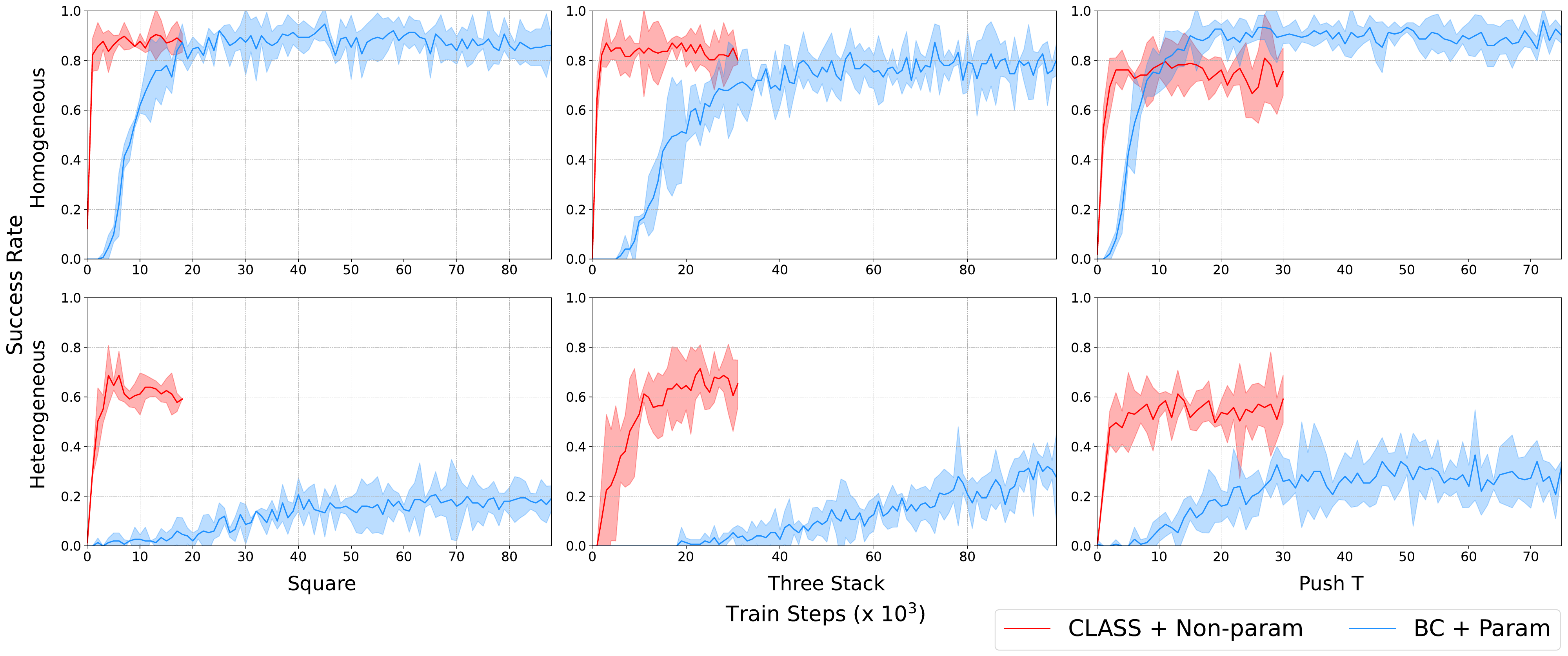}
\caption{Success Rate during Training for Dynamic Cam Square, Dynamic Cam Three Stack, and Random Color Push-T Rasks. In this plot, CLASS utilizes non-parametric rollouts using the learned representation, whereas BC uses diffusion policy head with parametric rollouts. CLASS requires significantly less number of training steps than BC and is more compute-efficient in repeated training runs.}
        \label{fig:train_speed}
    \end{center}
\end{figure*}

Although pre-computation introduces additional computational overheads, CLASS benefits from fast convergence during both pre-training and fine-tuning. CLASS pre-training converges much faster than BC training from scratch, as shown in Figure \ref{fig:train_speed} across different tasks. In addition, as shown in Table \ref{tab:bc_task_specific}, CLASS works with a fewer number of epochs during BC fine-tuning due to fast convergence rate for the policy learning, thanks to action-centric pre-training. This means that in repeated training runs, CLASS becomes more computationally efficient than BC, where there are multiple pre-training or fine-tuning attempts following a one-time pre-computation.

\section{Scaling Law}

\begin{figure*}[h]
    \begin{center}
        \includegraphics[width=0.8\linewidth]{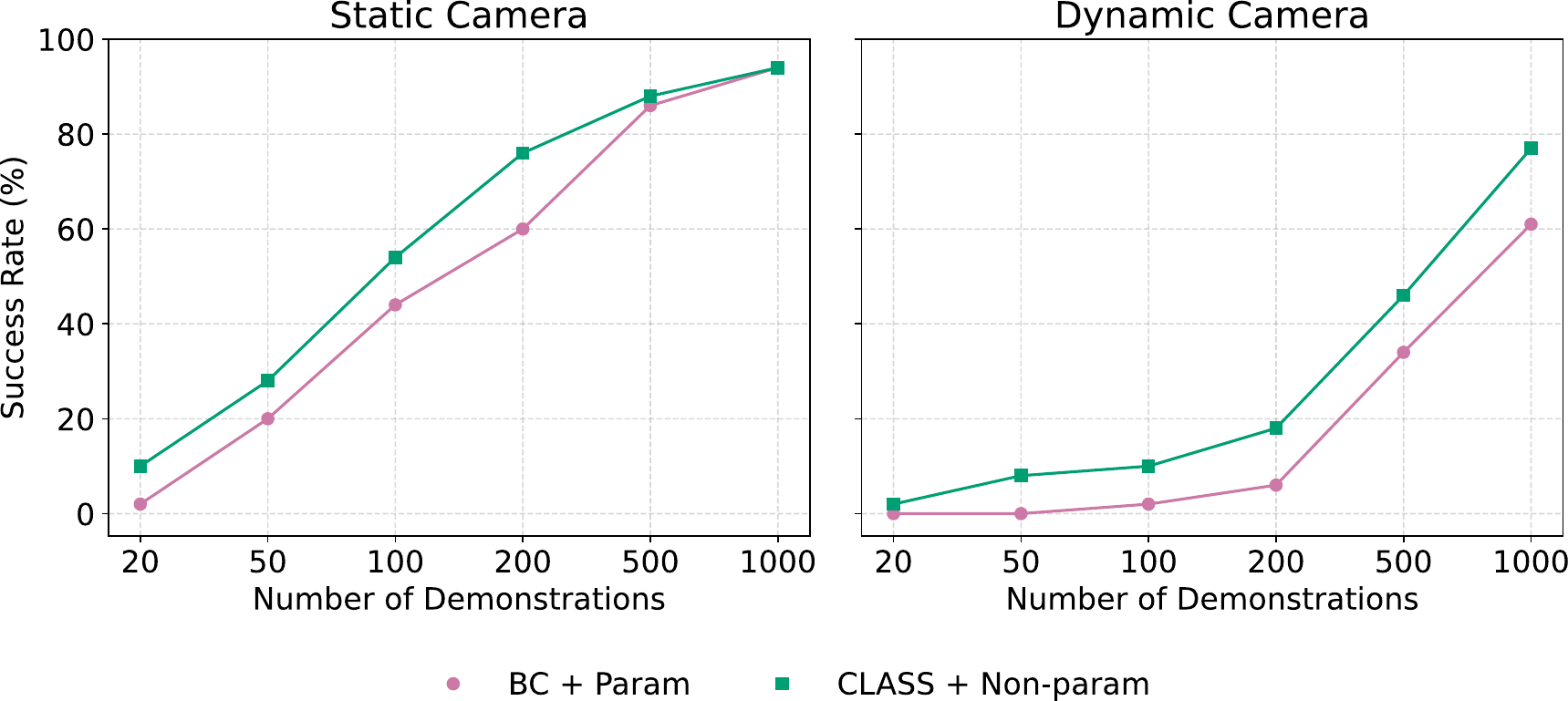}
\caption{Success rates on the Three-Stack task under varying dataset sizes (20–1000 demonstrations). We compare CLASS pre-training and behavioral cloning (BC) from scratch under both fixed and dynamic camera settings. CLASS consistently outperforms BC across all data regimes, demonstrating its scalability with larger datasets.}
        \label{fig:scaling_law}
    \end{center}
\end{figure*}

Similar to behavioral cloning (BC), CLASS follows a scaling law with improved performance as the dataset size increases. Larger datasets yield higher-quality and behaviorally-relevant positive pairs, thereby improving the contrastive signal and learned representations.
To illustrate this trend, we evaluate policy performance using 20, 50, 100, 200, 500, and 1000 demonstrations on the Three-Stack task, as shown in Fig.~\ref{fig:scaling_law}. CLASS with non-parametric rollouts consistently outperforms BC with parametric rollouts across different dataset sizes for both fixed and dynamic camera settings.
\vspace{4pt}

\vspace{4pt}

\section{Inference Time}

\begin{figure*}[h]
    \begin{center}
        \includegraphics[width=0.99\linewidth]{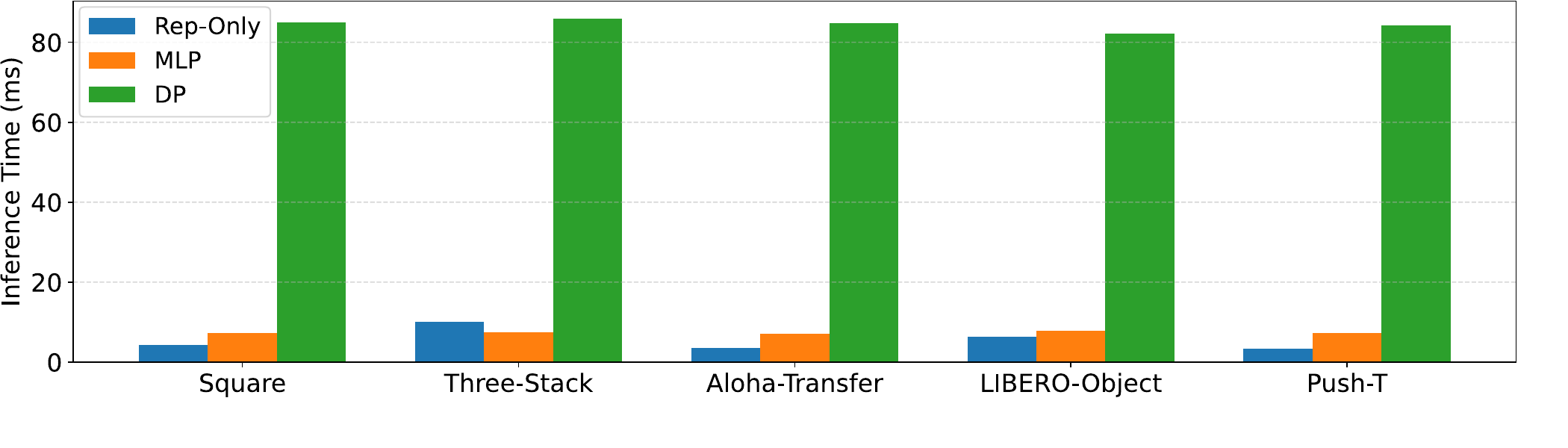}
        \vspace{-14pt}
\caption{
Inference time across policy classes and simulation tasks, averaged over 100 runs on an NVIDIA A40 GPU. Rep-Only methods are fastest on average but scale with dataset size due to the retrieval process. MLP and DP have fixed inference costs, with DP being slower due to its iterative denoising process. The dataset sizes for Square, Three-Stack, Aloha-Transfer, LIBERO-Object, and Push-T are 30K, 254K, 40K, 75K, and 27K samples, respectively.
}
        \label{fig:visualize_rep}
    \end{center}
\label{fig:inference}
\end{figure*}

Inference efficiency varies across policy classes. Representation-only methods avoid the overhead of a policy head and rely solely on embedding retrieval, resulting in fast inference, especially for small datasets. This makes them well-suited for high-frequency control. However, their inference time scales with dataset size due to retrieval costs. We compare inference times across policy classes to show these differences in Figure \ref{fig:inference}. On average, Rep-Only methods achieve an inference time of 5.5 ms, compared to 7.3 ms for MLP and 84.4 ms for DP.

\vspace{-14pt}
\section{Visualization of Representation}
\subsection{Latent Trajectory}
We visualize latent trajectories for the Dynamic Camera Square task in Fig.~\ref{fig:visualize_rep} to assess representation quality. Models trained with standard behavioral cloning and with CLASS pre-training are compared. CLASS learns representations where different demonstrations align and overlap in latent space, indicating higher transferability. In contrast, the BC-trained representation shows little overlap across demonstrations, suggesting poor generalization and less structured embeddings.

\begin{figure*}[h]
    \begin{center}
        \includegraphics[width=0.9\linewidth]{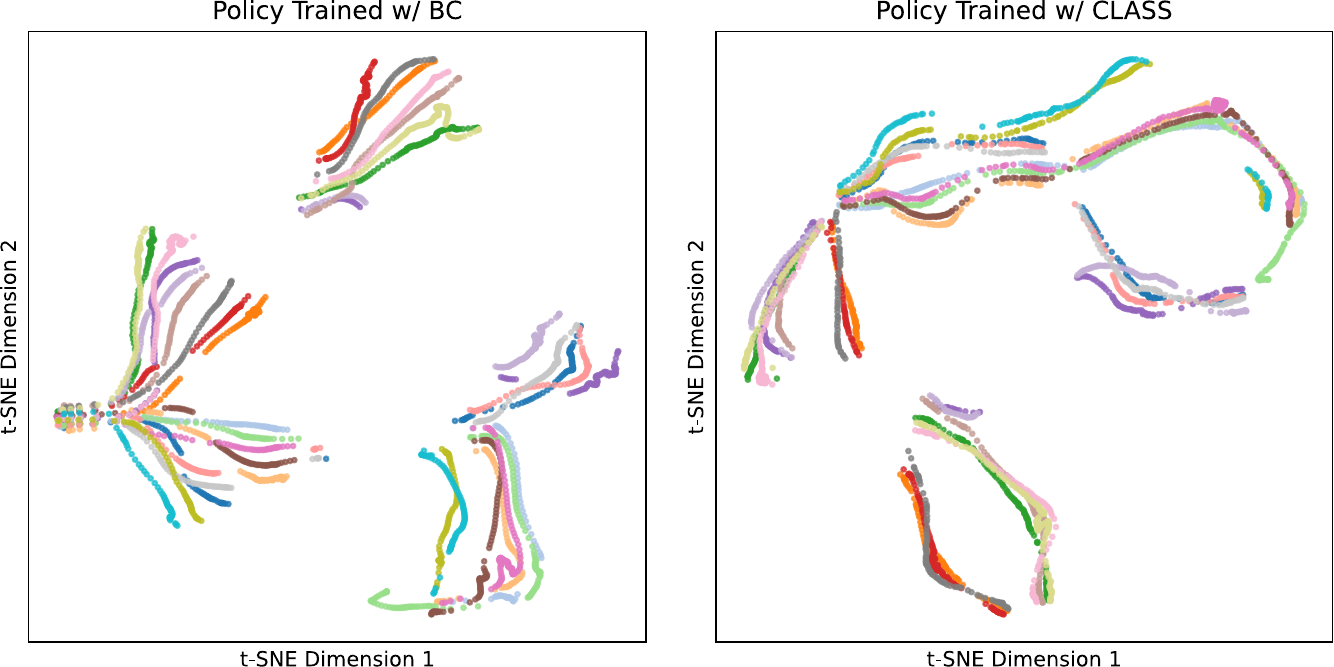}
\caption{Visualization of latent representations for the Dynamic Camera Square task using diffusion policy trained with (left) a standard behavioral cloning objective from scratch and (right) CLASS pre-training. Each colored spline represents an individual episode, with 20 episodes shown per plot. The embeddings are projected using t-SNE with a perplexity of 100.}

        \label{fig:rep_visualization}
    \end{center}
\end{figure*}

\subsection{Nearest Neighbors}

\begin{figure*}[h]
    \begin{center}
        \includegraphics[width=1.\linewidth]{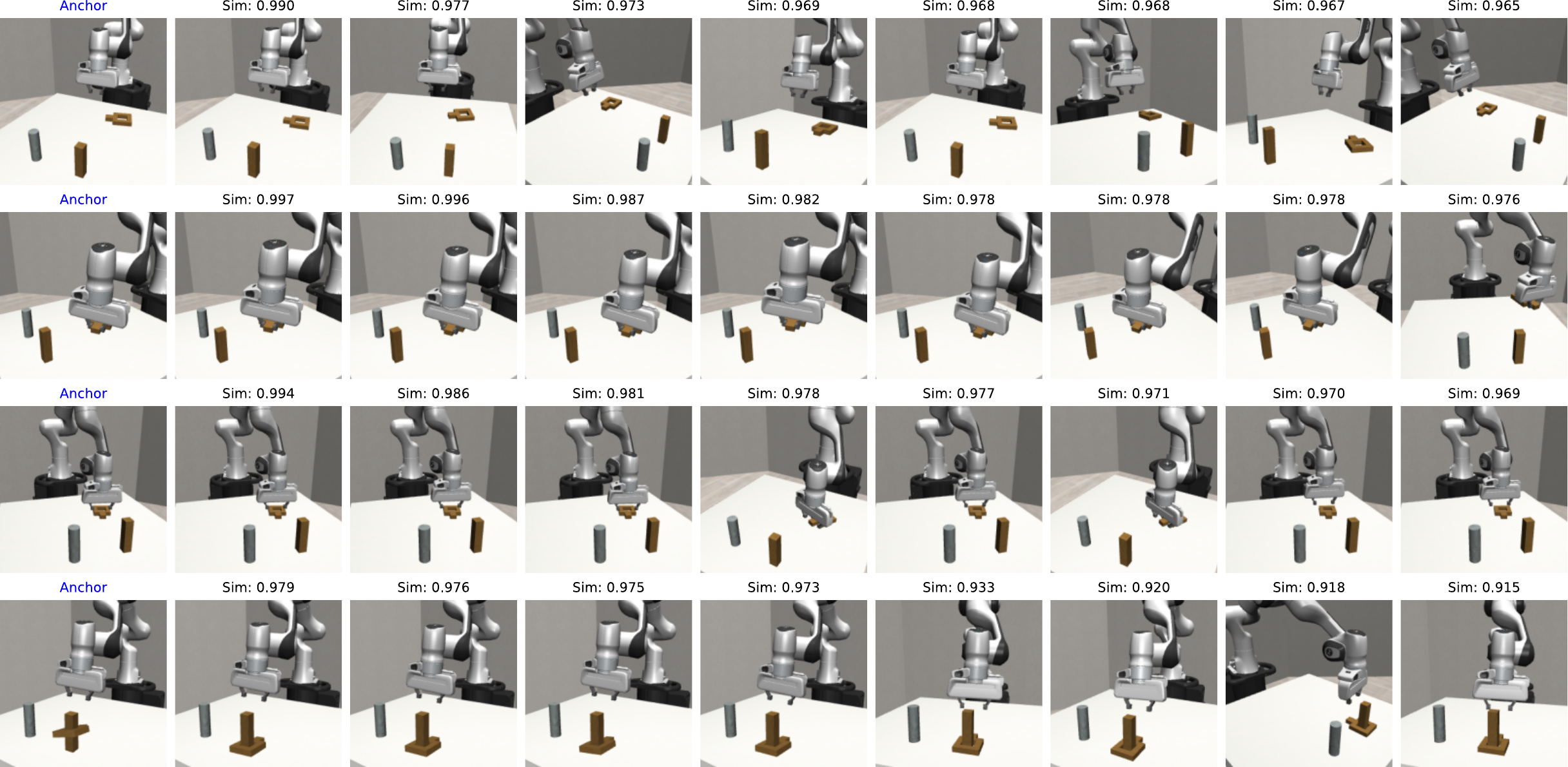}
\caption{Visualizations of top 8 nearest neighbors for Dynamic Camera Square task with \textbf{CLASS} pre-trained policy. The nearest neighbors are retrieved using cosine similarity in the latent space. It can be seen that similar environment scenes with varying viewpoints are identified as closest neighbors.}
        \label{fig:neareast_neighbor_CLASS}
    \end{center}
\end{figure*}

\begin{figure*}[h]
    \begin{center}
        \includegraphics[width=1.\linewidth]{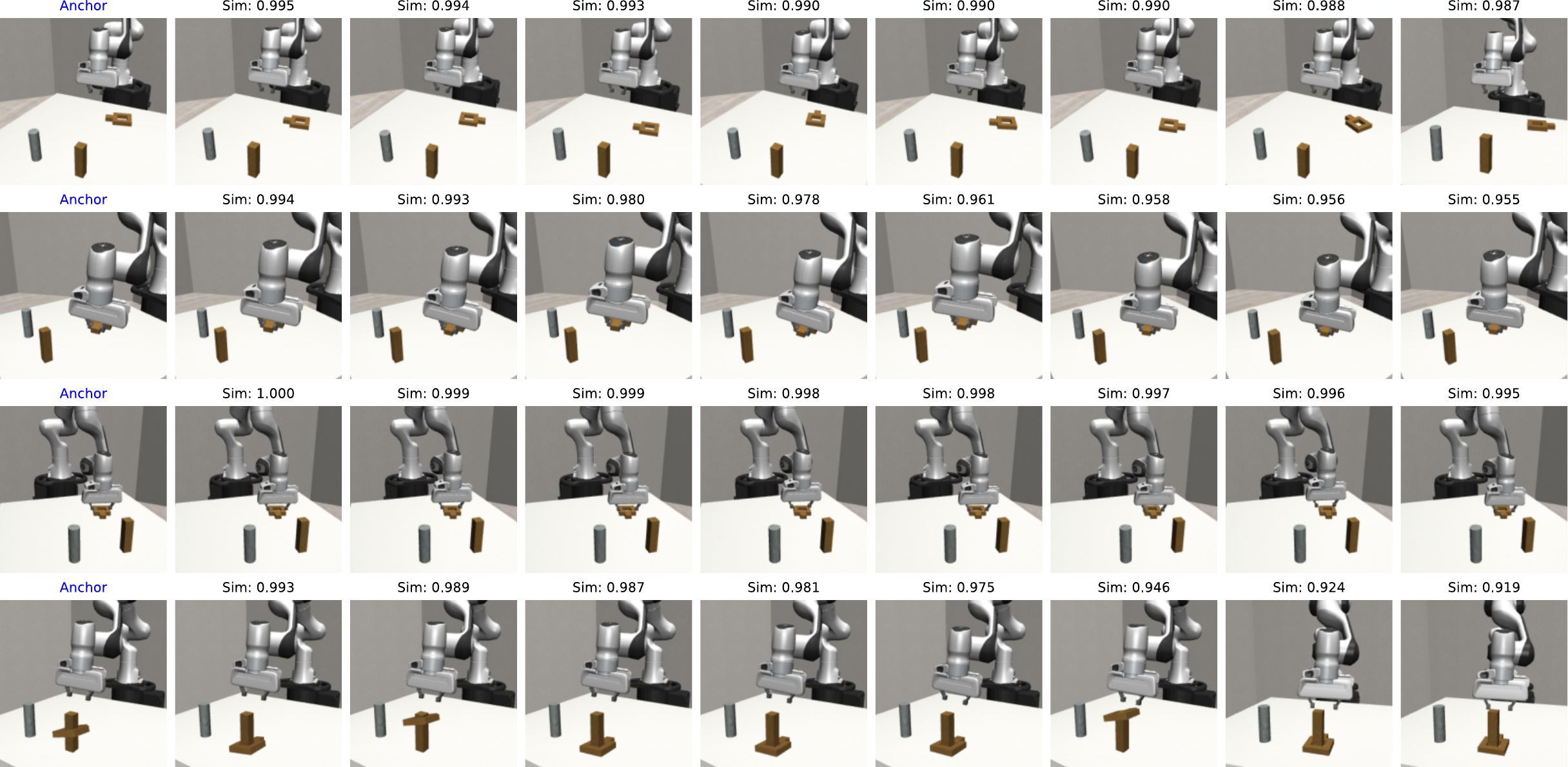}
\caption{Visualizations of top 8 nearest neighbors for Dynamic Camera Square task with \textbf{BC} trained diffusion policy with ImageNet pre-trained weights. Unlike CLASS pre-trained policy, nearest neighbors exhibit little variations in viewpoints.}
        \label{fig:neareast_neighbor_BC}
    \end{center}
\end{figure*}

We additionally visualize nearest neighbors using CLASS-trained representations in Figure ~\ref{fig:neareast_neighbor_CLASS} and compare to BC-trained representations in Figure \ref{fig:neareast_neighbor_BC}. Specifically, for Square task with varying viewpoints, it is observed that CLASS allows the representation from different viewpoints to be clustered if the future expert action sequences match, whereas BC exhibits little variations in viewpoints between neighbors.

\section{Real Robot Experiments}
\label{appx:realrobot}

We conduct real-world experiments using a Franka Research 3 robot with the UMI gripper \citep{chi2024universal}. Visual observations are captured by an Intel RealSense D435 camera mounted on a tripod, recording RGB images at a resolution of \(1280 \times 720\). These images are subsequently cropped and resized to \(256 \times 256 \times 3\). All experiments are executed on a single workstation equipped with an NVIDIA RTX 4090 GPU. Detailed descriptions of initialization, subtasks, and task completion criteria are provided below.

\textbf{Two-Stack}  
\begin{itemize}
    \item \textbf{Initialized:} The red and green cubes are randomly placed within a predefined \(30\,\text{cm} \times 15\,\text{cm}\) area.
    \item \textbf{Grasped:} The robot successfully picks up the red cube.
    \item \textbf{Stacked:} The red cube is placed stably on top of the green cube.
\end{itemize}

\textbf{Mug-Hang}  
\begin{itemize}
    \item \textbf{Initialized:} A mug is placed at a random pose on the table within a predefined \(10\,\text{cm} \times 10\,\text{cm}\) bounding box, with the handle visible from the agent's view.
    \item \textbf{Grasped:} The robot successfully grasps the mug.
    \item \textbf{Hung:} The mug is hung by its handle on a horizontal rod of the mug tree.

\end{itemize}

\textbf{Toaster-Load}  
\begin{itemize}
    \item \textbf{Initialized:} A mock soft bread slice is placed vertically using a support handle. The toaster is placed at a random orientation with a pre-defined position.
    \item \textbf{Grasped:} The robot successfully lifts the bread slice from the holder.
    \item \textbf{Loaded:} The bread is fully inserted into the toaster slot.
\end{itemize}

\begin{figure*}[h]
    \begin{center}
        \includegraphics[width=1.0\linewidth]{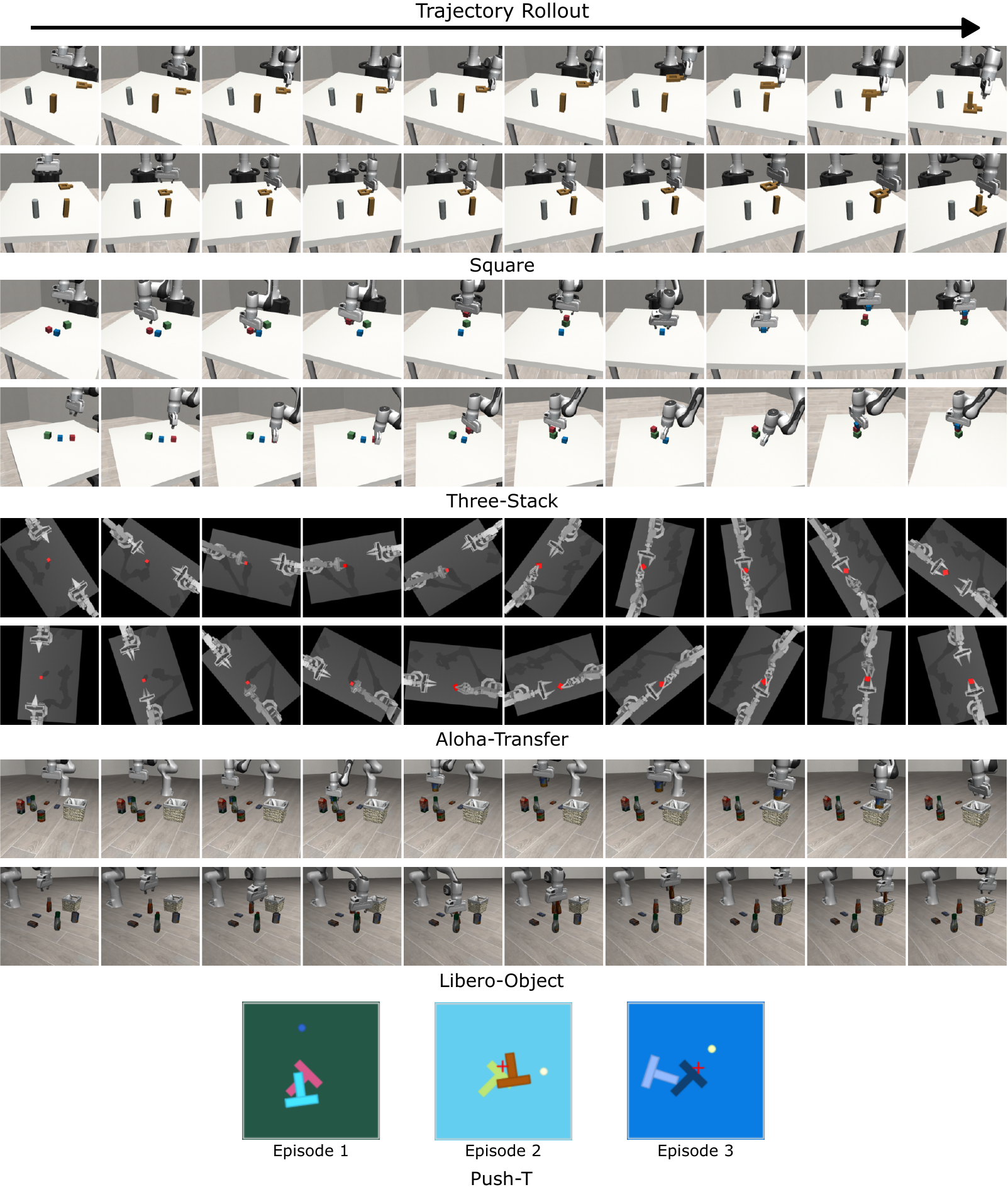}
\caption{Visualizations of heterogeneous episodes in the simulation tasks. For all tasks except Push-T, trajectories progress from left to right.}
        \label{fig:rollout}
    \end{center}
\end{figure*}

\begin{figure*}[h]
    \begin{center}
        \includegraphics[width=1.05\linewidth]{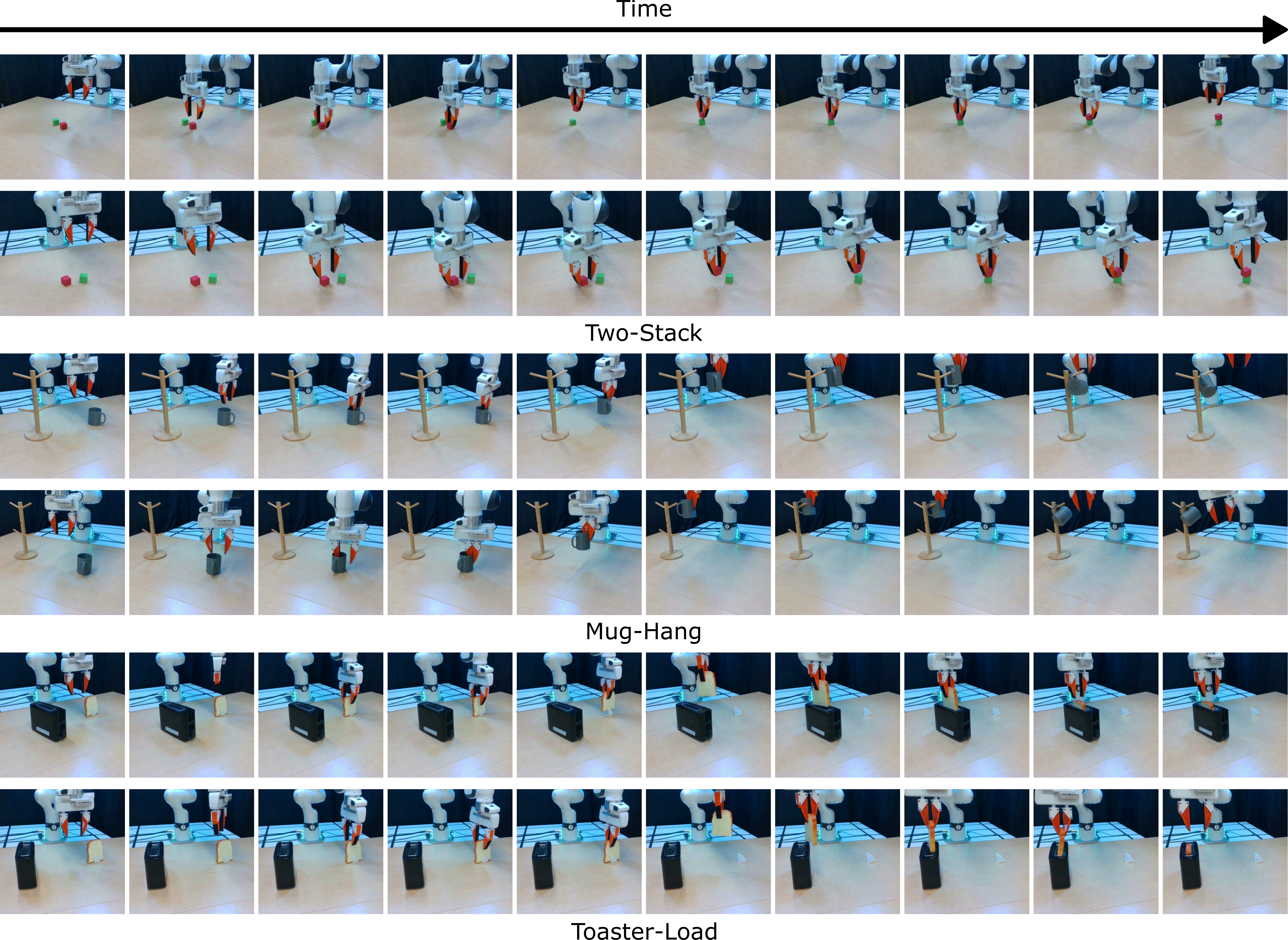}
\caption{Visualizations of successful evaluation rollouts for real-world tasks. Before each episode, the camera is randomly placed to introduce the visual shifts.}
        \label{fig:rollout}
    \end{center}
\end{figure*}

\end{document}